\pdfoutput=1

\documentclass[11pt]{article}

\usepackage[]{ACL2023}

\usepackage{times}
\usepackage{latexsym}
\usepackage{comment}
\usepackage{siunitx}
\newcommand*\sq{\mathbin{\vcenter{\hbox{\rule{.4ex}{.4ex}}}}}

\usepackage[T1]{fontenc}

\usepackage[utf8]{inputenc}

\usepackage{microtype}

\usepackage{inconsolata}

\usepackage{graphicx}
\usepackage{multirow}
\usepackage{booktabs}
\usepackage{adjustbox}
\usepackage{amsmath}
\usepackage{xcolor}  
\definecolor{labeled}{rgb}{0.98, 0.92, 0.88}
\definecolor{unlabeled}{rgb}{0.88, 0.96, 1.0}
\definecolor{kellygreen}{rgb}{0.3, 0.73, 0.09}
\definecolor{u0}{rgb}{0.97, 0.90, 0.85}
\definecolor{u10}{rgb}{0.85, 0.94, 0.98}
\definecolor{u100}{rgb}{0.92, 0.89, 0.99}

\title{Distill or Annotate? \\ Cost-Efficient Fine-Tuning of Compact Models}

\author{Junmo Kang, Wei Xu, Alan Ritter \\
  Georgia Institute of Technology \\
 \texttt{junmo.kang@gatech.edu} \\
 \texttt{\{wei.xu, alan.ritter\}@cc.gatech.edu} \\
 \\
 }

\begin{document}
\maketitle
\begin{abstract}
Fine-tuning large models is highly effective, however, inference can be expensive and produces carbon emissions. Knowledge distillation has been shown to be a practical solution to reduce inference costs, but the distillation process itself requires significant computational resources.
Rather than buying or renting GPUs to fine-tune, then distill a large model, an NLP practitioner might instead choose to allocate the available budget to hire annotators and manually label additional fine-tuning data.
In this paper, we investigate how to most efficiently use a fixed budget to build a compact model.
Through extensive experiments on six diverse tasks, we show that distilling from \texttt{T5-XXL} (11B) to \texttt{T5-Small} (60M) is almost always a cost-efficient strategy compared to annotating more data to directly train a compact model (\texttt{T5-Small}). We further investigate how the optimal budget allocated towards computation varies across scenarios.
We will make our code, datasets, annotation cost estimates, and baseline models available as a benchmark to support further work on cost-efficient training of compact models.

\end{abstract}

\section{Introduction}
Increasing the size of pre-trained models can consistently improve performance on downstream tasks after fine-tuning, as seen in studies based on BERT \citep{devlin2019bert}, RoBERTa \citep{liu2019roberta}, BART \citep{lewis2019bart}, T5 \citep{2020t5}, and the work on empirical scaling laws \citep{NEURIPS2020_1457c0d6,lester2021power,hernandez2021scaling}. However, using large models for inference is expensive and contributes to carbon emissions \citep{patterson2021carbon}. To address this, researchers have explored methods to compress large models through techniques such as knowledge distillation \citep{44873,sanh2019distilbert,gou2021knowledge}, which is effective in reducing inference costs \cite{magister2022teaching} and improving the generalization of smaller student models \cite{stanton2021does}. Nonetheless, the distillation process itself still requires significant computational, memory, and storage resources \citep{xia-etal-2022-structured}.

\begin{figure}[t]
    \centering
    \includegraphics[scale=0.33]{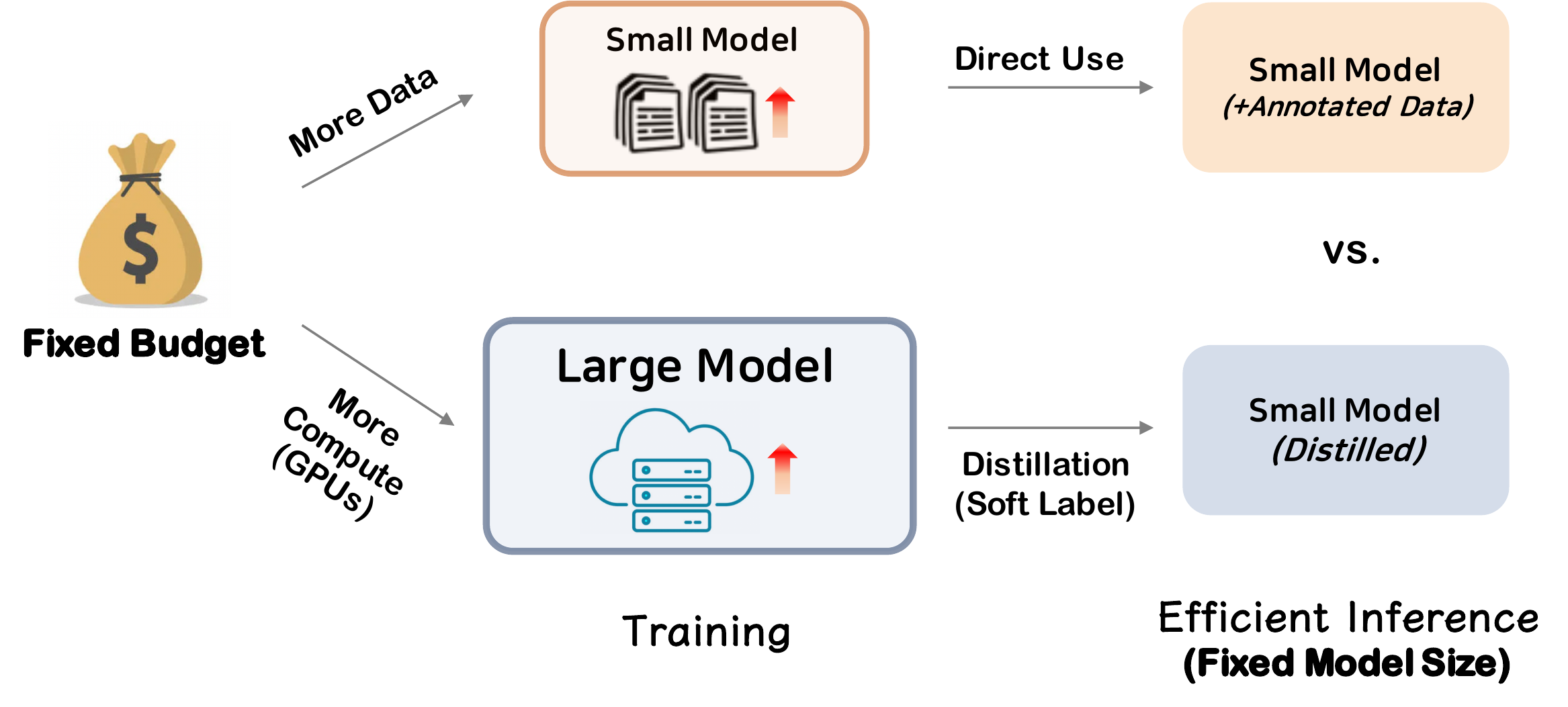}
    \caption{An illustration of two practical strategies to build a compact fixed-size model. Given a fixed budget and a small amount of initially annotated data, (i) one can annotate more data to directly fine-tune a small model. (ii) Alternatively, one may leverage a larger model with more computational resources to distill its knowledge into a small model for efficient inference.}
    \label{fig:strategies}
\end{figure}

In addition to compressing models, an alternative approach to improve performance without increasing inference costs is to simply label additional data for fine-tuning. Recent work has shown that a few hundred extra labels can sometimes lead to better performance than billions of additional model parameters \citep{Kirstain2021AFM}. This raises the question of how to most efficiently use a fixed budget to train a compact model which supports efficient inference while maximizing performance. 
One option is to use an available budget to hire annotators to label additional data and directly fine-tune a small model. Alternatively, the budget could be used to purchase or rent GPUs to fine-tune and distill a large teacher model (see Figure \ref{fig:strategies}).

\begin{table*}[th!]
\centering
\begin{adjustbox}{width=0.95\textwidth}
\begin{tabular}{llS[output-decimal-marker={,}]SS[output-decimal-marker={,}]}
\toprule
\textbf{Dataset} & \textbf{Task} & \textbf{\#Train} & \textbf{\$/Label} & \textbf{Total \$} \\
\midrule\midrule
\textbf{WLP} \footnotesize{\cite{tabassum-etal-2020-wnut}} & Named Entity Recognition & 11,966 & \$0.260 & \$3,111 \\
\textbf{\textsc{Stanceosaurus}} \footnotesize{\cite{Zheng2022StanceosaurusCS}} & Stance Classification & 12,130 & \$0.364 & \$4,415 \\
\textbf{FEVER} \footnotesize{\cite{thorne-etal-2018-fever}} & Fact Verification & 104,966 & \$0.129 & \$13,544 \\
\textbf{\textsc{MultiPIT$_\texttt{Id}$}} \footnotesize{\cite{Dou2022ImprovingLP}} & Paraphrase Identification & 92,217 & \$0.200 & \$18,443 \\
\textbf{\textsc{MultiPIT$_\texttt{Gen}$}} \footnotesize{\cite{Dou2022ImprovingLP}} & Paraphrase Generation & 49,673 & \$0.371 & \$18,443 \\
\textbf{\textsc{Natural Questions}} \footnotesize{\cite{kwiatkowski-etal-2019-natural}} & Question Answering & 87,372 & \$0.129 &	\$11,271 \\
\bottomrule
\end{tabular}
\end{adjustbox}
\caption{Data annotation costs for various NLP datasets/tasks.
}
\label{Table:cost_analysis}
\end{table*}

In this paper, we use the theory of consumer choice \cite{10.2307/2228949, 10.2307/1828835,bai-etal-2021-pre} to investigate the question of 
when distillation is a cost-efficient strategy for model compression. 
Based on extensive empirical analysis, we provide recommendations on how to allocate a fixed budget for human annotation and computing resources to train a compact model. Our experiments across six NLP tasks reveal that distillation with unlabeled data is almost always a cost-efficient strategy for improving the performance of compact models when compared to annotation (see Table \ref{Table:main}). Furthermore, our analysis shows that the optimal allocation of budget towards distillation increases as more labeled data becomes available (see \S \ref{sec:initial_data} and Figure \ref{fig:pareto}). For smaller budgets, it is Pareto optimal \citep{abdolrashidi2021pareto,treviso2022predicting} to use smaller amounts of unlabeled data for distillation, while increasing the amount of labeled data, as this leads to a more knowledgeable teacher. As the budget increases, it becomes economical to distill using larger unlabeled datasets, because the teacher model outperforms the student by a significant margin.
Finally, we investigate the cost efficiency of data annotation with \texttt{GPT-3.5} \citep{ouyang2022training} (Figure \ref{fig:gpt-3.5}). We find that, although \texttt{GPT-3.5} is cheaper than human annotators, fine-tuning \texttt{T5-XXL} and then distilling a small model is more cost-efficient than directly fine-tuning the small model with pseudo-labels from \texttt{GPT-3.5}.

We will make our code, datasets, annotation cost estimates, and baseline models available as a benchmark to support further work on cost-efficient training of compact models.

\section{Study Design}
In this section, we first describe how we formulate the problem for the cost-efficiency study (\S \ref{problem_formulation}). We then compare two strategies (\S \ref{str1} \& \ref{str2}) for building compact models that incur different proportions of computational and human annotation costs. Finally, we explain how to estimate the annotation cost (\S \ref{annotation_cost_est}) and computational cost (\S \ref{computational_cost_est}) involved in the two strategies.

\subsection{Problem Formulation and Assumptions}
\label{problem_formulation}
The main focus of this study is to fairly evaluate the two approaches (\S \ref{str1} \& \S \ref{str2}) under a fixed budget. When financial constraints are in place, practitioners may be faced with weighing options of allocating money towards \textit{data} or \textit{compute}; we empirically investigate their trade-offs to maximize the resulting utility.
To enable extensive studies, we simulate the process of labeling data using a variety of existing crowdsourced datasets, and the cloud GPU rentals that charge per hour of use.


We assume the NLP engineer's salary is a fixed cost, so their time spent building models and/or managing a group of annotators to label data are not a factor in determining the total cost.  The only costs considered are the direct costs for human data labeling and GPU computation.
No task-specific labeled data is initially assumed to be available for free, but we do assume that pre-trained models such as \texttt{T5} \cite{2020t5}, which are publicly available, have zero cost.

\subsection{Strategy 1: Building a Compact Model Directly with Annotations \texttt{(Ann.)}}
\label{str1}
This strategy directly fine-tunes a compact model (e.g., \texttt{T5-Small} (60M)), allocating the entire budget towards human annotation. This is considered the most straightforward approach practitioners would choose to train a compact model.

In particular, given a budget constraint, we prepare data that can be maximally annotated using the budget, and we train \texttt{T5} \cite{2020t5} on the data under a unified text-to-text framework for all tasks (Table \ref{Table:cost_analysis}), maximizing the likelihood of a target text $Y$ given an input text $X$.
The format for an input $X$ and the corresponding target $Y$ for each task is detailed in Appendix \ref{input-output}.

Note that the most dominant cost associated with this strategy is the annotation cost. While the total cost of building this direct model can include the fine-tuning cost (i.e., computational cost), we found it negligible in most cases and thus omitted it, unless otherwise noted, for the sake of simplicity.\footnote{Fine-tuning \texttt{T5-Small} (60M) on 5K data, for example, takes less than half an hour, which costs approximately \$1, based on the computational cost in \S \ref{computational_cost_est}.}

\subsection{Strategy 2: Distilling from a Larger Model \texttt{(Dist.)}}
\label{str2}
As an alternative to annotating more data, one could allocate part of the budget towards computation to train a larger (e.g., \texttt{T5-XXL} (11B)) model on a smaller amount of data.  The large model can then be distilled to produce a final compact model that also supports efficient inference.

Following recent work \citep{xia-etal-2022-structured,zhou2022bert}, our study mostly focuses on task-specific model compression rather than general distillation \cite{sanh2019distilbert},\footnote{In general distillation, a pre-trained model is distilled before fine-tuning, such as DistillBERT.} however we provide analysis of general vs. task-specific distillation in Appendix \ref{additional_results}. General distillation requires significant computational resources; also task-specific and general distillation can be used together in a complementary fashion \citep{jiao2020tinybert}.

Notably, even for Strategy 2, annotated data is needed to train the large teacher model. Therefore, we assume to have a certain number ($N$) of data initially annotated by spending some part of the budget, and fine-tune the larger model using this data in the same way as in \S \ref{str1}. After that, a small model (i.e., student) is trained by distilling the larger model's (i.e., teacher) knowledge \cite{44873}, in which the teacher's probability distributions over a target sequence given a source input are used as soft labels. 
We adopt KL divergence loss, which compares two distributions, to make the student's distribution $P_{S}$ follow the teacher's output distribution $P_{T}$ with respect to task-specific unlabeled data\footnote{For example, source sentences without target paraphrased sentences for a paraphrase generation task. Refer to Appendix \ref{unlabeled_data} for details of the unlabeled data.}:
\begin{equation}
D_{KL}(P_{T}||P_{S}) = \sum_{v \in V} P_{T}(v) \log \frac{P_{T}(v)}{P_{S}(v)}
\end{equation}
where $V$ is vocabulary space. Input and target tokens that are conditioned to produce probabilities are omitted above for brevity.

The total cost includes both the initial cost for $N$ (the number of initially annotated training examples) and the computational cost for fine-tuning a large model and then distilling it into a compact model.

\begin{table*}[ht!]
\centering
\begin{adjustbox}{width=1\textwidth}
\begin{tabular}{lccrlrlrlrlrlrl}
\toprule
\multirow{4}{*}{\textbf{Task}} & \multirow{4}{*}{\textbf{\textit{\colorbox{labeled}{$N$}}} (\texttt{Initial \$})} & \multirow{4}{*}{\textbf{Strategy}}
& \multicolumn{10}{c}{\textbf{\texttt{Additional \$}}} \\
\cmidrule(lr){7-11}
    & & & \multicolumn{10}{c}{\texttt{Ann.} Performance \quad \footnotesize{(\textit{\colorbox{labeled}{\#Additional Data}})}} \\
    & & & \multicolumn{10}{c}{\texttt{Dist.} Performance \quad \footnotesize{(\textit{\colorbox{unlabeled}{GPU Hours / \#Unlabeled Data}})}} \\
\midrule\midrule

\multirow{7}{*}{\textbf{WLP}} & \multirow{4.5}{*}{\textit{\colorbox{labeled}{1K}} (\texttt{\$260})}
 & & \multicolumn{2}{c}{\texttt{+\$0}} & \multicolumn{2}{c}{\texttt{+\$100}} & \multicolumn{2}{c}{\texttt{+\$200}} & \multicolumn{2}{c}{\texttt{+\$300}} & \multicolumn{2}{c}{\texttt{+\$500}} \\
\cmidrule(lr){4-5}\cmidrule(lr){6-7}\cmidrule(lr){8-9}\cmidrule(lr){10-11}\cmidrule(lr){12-13}
 & & \texttt{T5-Small (Ann.)} & 
 \textbf{40.7} & \footnotesize{(\textit{\colorbox{labeled}{+0}})} & 
 50.0 & \footnotesize{(\textit{\colorbox{labeled}{+384}})} & 
 53.7 & \footnotesize{(\textit{\colorbox{labeled}{+769}})} & 
 57.8 & \footnotesize{(\textit{\colorbox{labeled}{+1153}})} & 
 62.7 & \footnotesize{(\textit{\colorbox{labeled}{+1923}})} \\
 & & \texttt{T5-XXL [\textbf{72.4}] $\Rightarrow$ T5-Small (Dist.)
 } & 
 \multicolumn{2}{c}{\footnotesize{N/A}} &
 \textbf{71.1} & \footnotesize{(\textit{\colorbox{unlabeled}{54h / 19K}})} & 
 \textbf{71.3} & \footnotesize{(\textit{\colorbox{unlabeled}{107h / 42K}})} &
 \textbf{70.9} & \footnotesize{(\textit{\colorbox{unlabeled}{160h / 65K}})} & 
 \textbf{70.8} & \footnotesize{(\textit{\colorbox{unlabeled}{267h / 111K}})} \\
 \cmidrule(lr){2-13}
 \multirow{7}{*}{} & \multirow{2}{*}{\textit{\colorbox{labeled}{5K}} (\texttt{\$1300})}
 & \texttt{T5-Small (Ann.)} & 
 \textbf{67.4} & \footnotesize{(\textit{\colorbox{labeled}{+0}})} & 
 \textbf{68.2} & \footnotesize{(\textit{\colorbox{labeled}{+384}})} & 
 68.6 & \footnotesize{(\textit{\colorbox{labeled}{+769}})} & 
 68.7 & \footnotesize{(\textit{\colorbox{labeled}{+1153}})} & 
 69.3 & \footnotesize{(\textit{\colorbox{labeled}{+1923}})} \\
 & & \texttt{T5-XXL [\textbf{74.2}] $\Rightarrow$ T5-Small (Dist.)} & 
 \multicolumn{2}{c}{\footnotesize{N/A}} &
 65.3 & \footnotesize{(\textit{\colorbox{unlabeled}{54h / 7K}})} & 
 \textbf{71.8} & \footnotesize{(\textit{\colorbox{unlabeled}{107h / 30K}})} &
 \textbf{72.4} & \footnotesize{(\textit{\colorbox{unlabeled}{160h / 53K}})} & 
 \textbf{72.5} & \footnotesize{(\textit{\colorbox{unlabeled}{267h / 99K}})} \\
\midrule\midrule

\multirow{6.5}{*}{\textbf{\textsc{Stanceo-}}} & \multirow{4.5}{*}{\textit{\colorbox{labeled}{1K}} (\texttt{\$364})}
 & & \multicolumn{2}{c}{\texttt{+\$0}} & \multicolumn{2}{c}{\texttt{+\$100}} & \multicolumn{2}{c}{\texttt{+\$150}} & \multicolumn{2}{c}{\texttt{+\$200}} & \multicolumn{2}{c}{\texttt{+\$300}} \\
\cmidrule(lr){4-5}\cmidrule(lr){6-7}\cmidrule(lr){8-9}\cmidrule(lr){10-11}\cmidrule(lr){12-13}
 & & \texttt{T5-Small (Ann.)} & 
 \textbf{37.5} & \footnotesize{(\textit{\colorbox{labeled}{+0}})} & 
 45.4 & \footnotesize{(\textit{\colorbox{labeled}{+274}})} & 
 45.5 & \footnotesize{(\textit{\colorbox{labeled}{+412}})} & 
 45.5 & \footnotesize{(\textit{\colorbox{labeled}{+549}})} & 
 44.7 &  \footnotesize{(\textit{\colorbox{labeled}{+824}})} \\
 & & \texttt{T5-XXL [\textbf{62.5}] $\Rightarrow$ T5-Small (Dist.)
 } & 
 \multicolumn{2}{c}{\footnotesize{N/A}} &
 \textbf{54.2} & \footnotesize{(\textit{\colorbox{unlabeled}{54h / 37K}})} & 
 \textbf{54.6} & \footnotesize{(\textit{\colorbox{unlabeled}{80h / 60K}})} &
 \textbf{56.3} & \footnotesize{(\textit{\colorbox{unlabeled}{107h / 82K}})} & 
 \textbf{56.9} & \footnotesize{(\textit{\colorbox{unlabeled}{160h / 126K}})} \\
\cmidrule(lr){2-13}
 \multirow{-1.5}{*}{\textbf{\textsc{saurus}} } & \multirow{2}{*}{\textit{\colorbox{labeled}{5K} (\texttt{\$1820})}}
 & \texttt{T5-Small (Ann.)} & 
 \textbf{49.4} & \footnotesize{(\textit{\colorbox{labeled}{+0}})} & 
 50.7 & \footnotesize{(\textit{\colorbox{labeled}{+274}})} & 
 52.6 &  \footnotesize{(\textit{\colorbox{labeled}{+412}})} & 
 49.1 & \footnotesize{(\textit{\colorbox{labeled}{+549}})} &
 50.3 &  \footnotesize{(\textit{\colorbox{labeled}{+824}})} \\
 & & \texttt{T5-XXL [\textbf{69.6}] $\Rightarrow$ T5-Small (Dist.)} & 
 \multicolumn{2}{c}{\footnotesize{N/A}} & 
 \textbf{52.4} & \footnotesize{(\textit{\colorbox{unlabeled}{54h / 17K}})} & 
 \textbf{55.4} & \footnotesize{(\textit{\colorbox{unlabeled}{80h / 40K}})} & \textbf{56.2} & \footnotesize{(\textit{\colorbox{unlabeled}{107h / 62K}})} & \textbf{60.5} & \footnotesize{(\textit{\colorbox{unlabeled}{160h / 106K}})} \\
\midrule\midrule

\multirow{7}{*}{\textbf{\textsc{FEVER}}} & \multirow{4.5}{*}{\textit{\colorbox{labeled}{1K}} (\texttt{\$129})}
 & & \multicolumn{2}{c}{\texttt{+\$0}} & \multicolumn{2}{c}{\texttt{+\$50}} & \multicolumn{2}{c}{\texttt{+\$75}} & \multicolumn{2}{c}{\texttt{+\$100}} & \multicolumn{2}{c}{\texttt{+\$150}} \\
\cmidrule(lr){4-5}\cmidrule(lr){6-7}\cmidrule(lr){8-9}\cmidrule(lr){10-11}\cmidrule(lr){12-13}
 & & \texttt{T5-Small (Ann.)} & 
 \textbf{49.7} & \footnotesize{(\textit{\colorbox{labeled}{+0}})} & 
 49.7 & \footnotesize{(\textit{\colorbox{labeled}{+387}})} & 
 49.7 & \footnotesize{(\textit{\colorbox{labeled}{+581}})} & 
 49.7 & \footnotesize{(\textit{\colorbox{labeled}{+775}})} & 
 49.8 &  \footnotesize{(\textit{\colorbox{labeled}{+1162}})} \\
 & & \texttt{T5-XXL [\textbf{73.5}] $\Rightarrow$ T5-Small (Dist.)
 } & 
 \multicolumn{2}{c}{\footnotesize{N/A}} &
 \textbf{71.3} & \footnotesize{(\textit{\colorbox{unlabeled}{27h / 54K}})} & 
 \textbf{71.1} & \footnotesize{(\textit{\colorbox{unlabeled}{40h / 86K}})} &
 \textbf{71.6} & \footnotesize{(\textit{\colorbox{unlabeled}{54h / 118K}})} & 
 \textbf{71.7} & \footnotesize{(\textit{\colorbox{unlabeled}{80h / 182K}})} \\
 \cmidrule(lr){2-13}
 \multirow{3}{*}{} & \multirow{2}{*}{\textit{\colorbox{labeled}{5K} (\texttt{\$645})}}
 & \texttt{T5-Small (Ann.)} & 
 \textbf{67.2} & \footnotesize{(\textit{\colorbox{labeled}{+0}})} & 
 68.2 & \footnotesize{(\textit{\colorbox{labeled}{+387}})} & 
 68.1 & \footnotesize{(\textit{\colorbox{labeled}{+581}})} & 
 68.1 &  \footnotesize{(\textit{\colorbox{labeled}{+775}})} & 
 68.9 & \footnotesize{(\textit{\colorbox{labeled}{+1162}})} \\
 & & \texttt{T5-XXL [\textbf{78.0}] $\Rightarrow$ T5-Small (Dist.)} & 
 \multicolumn{2}{c}{\footnotesize{N/A}} & 
 \textbf{73.4} & \footnotesize{(\textit{\colorbox{unlabeled}{27h / 35K}})} & \textbf{74.1} & \footnotesize{(\textit{\colorbox{unlabeled}{40h / 67K}})} & \textbf{74.3} & \footnotesize{(\textit{\colorbox{unlabeled}{54h / 99K}})} & \textbf{74.8} & \footnotesize{(\textit{\colorbox{unlabeled}{80h / 163K}})} \\
\midrule\midrule

\multirow{7}{*}{\textbf{\textsc{MultiPIT$_\texttt{Id}$}}} & \multirow{4.5}{*}{\textit{\colorbox{labeled}{1K}} (\texttt{\$200})}
 & & \multicolumn{2}{c}{\texttt{+\$0}} & \multicolumn{2}{c}{\texttt{+\$100}} & \multicolumn{2}{c}{\texttt{+\$150}} & \multicolumn{2}{c}{\texttt{+\$200}} & \multicolumn{2}{c}{\texttt{+\$300}} \\
\cmidrule(lr){4-5}\cmidrule(lr){6-7}\cmidrule(lr){8-9}\cmidrule(lr){10-11}\cmidrule(lr){12-13}
 & & \texttt{T5-Small (Ann.)} & 
 \textbf{53.0} & \footnotesize{(\textit{\colorbox{labeled}{+0}})} &
 53.1 & \footnotesize{(\textit{\colorbox{labeled}{+500}})} &
 53.1 & \footnotesize{(\textit{\colorbox{labeled}{+750}})} &
 54.6 & \footnotesize{(\textit{\colorbox{labeled}{+1000}})} &
 54.2 &  \footnotesize{(\textit{\colorbox{labeled}{+1500}})} \\
 & & \texttt{T5-XXL [\textbf{79.9}] $\Rightarrow$ T5-Small (Dist.)
 } & 
 \multicolumn{2}{c}{\footnotesize{N/A}} & 
 \textbf{79.1} & \footnotesize{(\textit{\colorbox{unlabeled}{54h / 75K}})} & 
 \textbf{78.3} & \footnotesize{(\textit{\colorbox{unlabeled}{80h / 115K}})} & 
 \textbf{78.8} & \footnotesize{(\textit{\colorbox{unlabeled}{107h / 156K}})} & 
 \textbf{77.9} & \footnotesize{(\textit{\colorbox{unlabeled}{160h / 237K}})} \\
 \cmidrule(lr){2-13}
 \multirow{3}{*}{} & \multirow{2}{*}{\textit{\colorbox{labeled}{5K} (\texttt{\$1000})}}
 & \texttt{T5-Small (Ann.)} & 
 \textbf{78.0} & \footnotesize{(\textit{\colorbox{labeled}{+0}})} & 
 77.4 & \footnotesize{(\textit{\colorbox{labeled}{+500}})} & 
 77.0 & \footnotesize{(\textit{\colorbox{labeled}{+750}})} & 
 78.1 &  \footnotesize{(\textit{\colorbox{labeled}{+1000}})} & 
 77.8 & \footnotesize{(\textit{\colorbox{labeled}{+1500}})} \\
 & & \texttt{T5-XXL [\textbf{84.5}] $\Rightarrow$ T5-Small (Dist.)} & 
 \multicolumn{2}{c}{\footnotesize{N/A}} & 
 \textbf{80.6} & \footnotesize{(\textit{\colorbox{unlabeled}{54h / 54K}})} & 
 \textbf{80.5} & \footnotesize{(\textit{\colorbox{unlabeled}{80h / 95K}})} & \textbf{81.1} & \footnotesize{(\textit{\colorbox{unlabeled}{107h / 136K}})} & \textbf{81.9} & \footnotesize{(\textit{\colorbox{unlabeled}{160h / 217K}})} \\
\midrule\midrule

\multirow{7}{*}{\textbf{\textsc{MultiPIT$_\texttt{Gen}$}}} & \multirow{4.5}{*}{\textit{\colorbox{labeled}{1K}} (\texttt{\$371})}
 & & \multicolumn{2}{c}{\texttt{+\$0}} & \multicolumn{2}{c}{\texttt{+\$100}} & \multicolumn{2}{c}{\texttt{+\$150}} & \multicolumn{2}{c}{\texttt{+\$200}} & \multicolumn{2}{c}{\texttt{+\$300}} \\
\cmidrule(lr){4-5}\cmidrule(lr){6-7}\cmidrule(lr){8-9}\cmidrule(lr){10-11}\cmidrule(lr){12-13}
 & & \texttt{T5-Small (Ann.)} & 
 \textbf{56.8} & \footnotesize{(\textit{\colorbox{labeled}{+0}})} & 
 57.7 & \footnotesize{(\textit{\colorbox{labeled}{+269}})} & 
 58.9 & \footnotesize{(\textit{\colorbox{labeled}{+404}})} & 
 59.2 & \footnotesize{(\textit{\colorbox{labeled}{+539}})} & 
 59.3 &  \footnotesize{(\textit{\colorbox{labeled}{+808}})} \\
 & & \texttt{T5-XXL [\textbf{67.4}] $\Rightarrow$ T5-Small (Dist.)
 } & 
 \multicolumn{2}{c}{\footnotesize{N/A}} & 
 \textbf{60.3} & \footnotesize{(\textit{\colorbox{unlabeled}{54h / 56K}})} & 
 \textbf{62.1} & \footnotesize{(\textit{\colorbox{unlabeled}{80h / 87K}})} & 
 \textbf{62.0} & \footnotesize{(\textit{\colorbox{unlabeled}{107h / 118K}})} & 
 \textbf{62.6} & \footnotesize{(\textit{\colorbox{unlabeled}{160h / 179K}})} \\
 \cmidrule(lr){2-13}
 \multirow{3}{*}{} & \multirow{2}{*}{\textit{\colorbox{labeled}{10K} (\texttt{\$3710})}}
 & \texttt{T5-Small (Ann.)} & 
 \textbf{68.6} & \footnotesize{(\textit{\colorbox{labeled}{+0}})} & 
 \textbf{68.6} & \footnotesize{(\textit{\colorbox{labeled}{+269}})} & 
 68.6 & \footnotesize{(\textit{\colorbox{labeled}{+404}})} & 
 68.6 &  \footnotesize{(\textit{\colorbox{labeled}{+539}})} & 
 68.7 & \footnotesize{(\textit{\colorbox{labeled}{+808}})} \\
 & & \texttt{T5-XXL [\textbf{74.8}] $\Rightarrow$ T5-Small (Dist.)} & 
 \multicolumn{2}{c}{\footnotesize{N/A}} & 
 68.4 & \footnotesize{(\textit{\colorbox{unlabeled}{54h / 10K}})} & 
 \textbf{72.1} & \footnotesize{(\textit{\colorbox{unlabeled}{80h / 41K}})} & \textbf{73.7} & \footnotesize{(\textit{\colorbox{unlabeled}{107h / 72K}})} & \textbf{74.0} & \footnotesize{(\textit{\colorbox{unlabeled}{160h / 133K}})} \\ 
\midrule\midrule

\multirow{6.5}{*}{\textbf{\textsc{Natural}}} & \multirow{4.5}{*}{\textit{\colorbox{labeled}{1K}} (\texttt{\$129})}
 & & \multicolumn{2}{c}{\texttt{+\$0}} & \multicolumn{2}{c}{\texttt{+\$50}} & \multicolumn{2}{c}{\texttt{+\$75}} & \multicolumn{2}{c}{\texttt{+\$100}} & \multicolumn{2}{c}{\texttt{+\$150}} \\
\cmidrule(lr){4-5}\cmidrule(lr){6-7}\cmidrule(lr){8-9}\cmidrule(lr){10-11}\cmidrule(lr){12-13}
 & & \texttt{T5-Small (Ann.)} & 
 \textbf{3.5} & \footnotesize{(\textit{\colorbox{labeled}{+0}})} & 
 4.1 & \footnotesize{(\textit{\colorbox{labeled}{+387}})} & 
 4.2 & \footnotesize{(\textit{\colorbox{labeled}{+581}})} & 
 4.5 & \footnotesize{(\textit{\colorbox{labeled}{+775}})} & 
 5.0 &  \footnotesize{(\textit{\colorbox{labeled}{+1162}})} \\
 & & \texttt{T5-XXL [\textbf{21.9}] $\Rightarrow$ T5-Small (Dist.)
 } & 
 \multicolumn{2}{c}{\footnotesize{N/A}} &
 \textbf{11.3} & \footnotesize{(\textit{\colorbox{unlabeled}{27h / 34K}})} & 
 \textbf{11.8} & \footnotesize{(\textit{\colorbox{unlabeled}{40h / 54K}})} &
 \textbf{13.0} & \footnotesize{(\textit{\colorbox{unlabeled}{54h / 75K}})} & 
 \textbf{13.5} & \footnotesize{(\textit{\colorbox{unlabeled}{80h / 115K}})} \\
 \cmidrule(lr){2-13}
\multirow{-1.5}{*}{\textbf{\textsc{Questions}}} & \multirow{2}{*}{\textit{\colorbox{labeled}{10K} (\texttt{\$1290})}}
 & \texttt{T5-Small (Ann.)} & 
 \textbf{9.8} & \footnotesize{(\textit{\colorbox{labeled}{+0}})} & 
 \textbf{10.2} & \footnotesize{(\textit{\colorbox{labeled}{+387}})} & 
 9.9 & \footnotesize{(\textit{\colorbox{labeled}{+581}})} & 
 10.4 &  \footnotesize{(\textit{\colorbox{labeled}{+775}})} & 
 10.3 & \footnotesize{(\textit{\colorbox{labeled}{+1162}})} \\
 & & \texttt{T5-XXL [\textbf{26.1}] $\Rightarrow$ T5-Small (Dist.)} & 
 \multicolumn{2}{c}{\footnotesize{N/A}} & 
 \multicolumn{2}{c}{\footnotesize{N/A}} & 
 \textbf{12.0} & \footnotesize{(\textit{\colorbox{unlabeled}{40h / 17K}})} & \textbf{16.3} & \footnotesize{(\textit{\colorbox{unlabeled}{54h / 46K}})} & \textbf{18.0} & \footnotesize{(\textit{\colorbox{unlabeled}{80h / 104K}})} \\

\bottomrule
\end{tabular}
\end{adjustbox}
\caption{Main results of the cost efficiency of a small model with more\texttt{ data annotation (Ann.)} and \texttt{teacher [\textbf{performance}] $\Rightarrow$ student distillation (Dist.)} on various NLP tasks. \colorbox{labeled}{$N$} indicates the number of starting data annotated with the corresponding (\texttt{initial \$}). (\textit{\colorbox{labeled}{\#Additional Data}}) refers to the number of annotated data additional to \colorbox{labeled}{$N$}, and (\textit{\colorbox{unlabeled}{GPU Hours}}) denotes the total GPU hours for fine-tuning the teacher model on \colorbox{labeled}{$N$} data, plus for the distillation into a small model using varied (\textit{\colorbox{unlabeled}{\#Unlabeled Data}}). N/A is used when it is not feasible to build a model given the cost. 
}
\label{Table:main}
\end{table*}

\subsection{Cost Estimation for Data Annotation}
\label{annotation_cost_est}
This study considers six diverse and practical NLP tasks, shown in Table \ref{Table:cost_analysis}. We estimate the annotation cost for each dataset based on mentions in the corresponding literature if available, correspondence with creators of the dataset, or prices of the Data Labeling Service from Google Cloud, following \citet{wang-etal-2021-want-reduce}\footnote{\url{https://cloud.google.com/ai-platform/data-labeling/pricing\#labeling_costs}}. 
Detailed descriptions of our cost estimates for each dataset are provided in Appendix \ref{ann-cost-details}.

\subsection{Estimation of Computational Cost}
\label{computational_cost_est}
This work assumes that computing resources are rented from Google Cloud for model training. We specifically consider NVIDIA A100 GPUs, each equipped with 40GB of VRAM, to fit a large model (e.g., 11B parameters) into them. The price of this, which includes a virtual machine and storage, is set at about \texttt{\$3.75} per 1 GPU hour. For extensive studies, we exploit our own resources, A40 GPUs that have been shown to be approximately 2x slower than A100 through benchmark results\footnote{\url{https://lambdalabs.com/blog/nvidia-rtx-a40-benchmarks}} as well as our preliminary experiment that compares the training time. As a result, we estimate the computational cost as \texttt{\$1.875} per 1 GPU hour. This is a realistic price that practitioners would need to pay, unlike theoretical measures such as FLOPs, which do not reflect the real runtime \cite{xu2022survey} and costs. An example breakdown of cost estimates for building compact models is provided in Appendix (Table \ref{Table:compute_cost_analysis}).

\section{Evaluating Annotation and Distillation under a Fixed Budget}
\label{settings}
In Table \ref{Table:main}, we evaluate the two strategies under varying budgets for six different tasks. We first set \colorbox{labeled}{\textit{N}}, the number of starting data annotated by spending an \texttt{initial \$}. Given a fixed budget, we then either \colorbox{labeled}{\textit{annotate more data}} for the \texttt{annotation} \texttt{(Ann.)} strategy, or use more \colorbox{unlabeled}{\textit{GPU hours}} along with more \colorbox{unlabeled}{\textit{unlabeled data}} for the \texttt{distillation (Dist.)} strategy. 

We consider \texttt{T5-Small} (60M) as a compact model and \texttt{T5-XXL} (11B) as a teacher model for our main study.
All models are fine-tuned based on \texttt{T5 v1.1} \cite{roberts-etal-2020-much}, which was pre-trained in an unsupervised way only, unlike the original \texttt{T5} \cite{2020t5}.

In the case of FEVER and \textsc{Natural Questions}, following \citet{lee-etal-2020-language} and \citet{roberts-etal-2020-much} respectively, we consider a closed-book setting where models should rely solely on its parametric knowledge, and report performances on dev sets as test sets are private. To measure performances, we use accuracy for FEVER and \textsc{MultiPIT$_\texttt{Id}$}, F1 for WLP, \textsc{Stanceosaurus}, and \textsc{Natural Questions}, and BERT-iBLEU \cite{niu-etal-2021-unsupervised} (i.e., the harmonic mean of self-BLEU and BERTS\textsc{core} \cite{Zhang*2020BERTScore:}) for \textsc{MultiPIT$_\texttt{Gen}$}. More details about experimental settings are described in Appendix \ref{hyperparameters}.

\begin{table*}[t]
\centering
\begin{adjustbox}{width=1\textwidth}
\begin{tabular}{lccccccc}
\toprule
\textbf{Model (\texttt{Teacher $\Rightarrow$ Student})} & \textbf{WLP} & \textbf{\textsc{Stanceosaurus}} & \textbf{FEVER} & \textbf{\textsc{MultiPIT$_\texttt{Id}$}} & \textbf{\textsc{MultiPIT$_\texttt{Gen}$}} & \textbf{\textsc{Natural Questions}} \\
\midrule\midrule
\texttt{T5-Small $\Rightarrow$ T5-Small (Self-Dist.)} & 65.2 [67.4] & 50.3 [50.5] & 67.6 [67.2] & 77.1 [78.0] & 66.1 [68.1] & 3.8 [9.8] \\
\texttt{T5-XXL $\Rightarrow$ T5-Small (Dist.)} & 70.6 [74.2] & 58.9 [69.6] & 74.2 [78.0] & 80.9 [84.5] & 73.8 [74.8] & 17.8 [26.1] \\
\bottomrule
\end{tabular}
\end{adjustbox}
\caption{Results of self-distillation and distillation with the same amount of unlabeled data (\textit{100K}). Numbers in [\ ] represent the performances of the teacher models that are trained on \textit{\textit{5K}} annotated data.}
\label{Table:analysis1_of_results}
\end{table*}

\begin{table*}[t]
\centering
\begin{adjustbox}{width=1\textwidth}
\begin{tabular}{lrlrlrlrlrlrl}
\toprule
\textbf{Model} & \multicolumn{2}{c}{\textbf{WLP}} & \multicolumn{2}{c}{\textbf{\textsc{Stanceosaurus}}} & \multicolumn{2}{c}{\textbf{FEVER}} & \multicolumn{2}{c}{\textbf{\textsc{MultiPIT$_\texttt{Id}$}}} & \multicolumn{2}{c}{\textbf{\textsc{MultiPIT$_\texttt{Gen}$}}} & \multicolumn{2}{c}{\textbf{\textsc{Natural Questions}}} \\
\midrule\midrule

\texttt{T5-XXL $\Rightarrow$ T5-Small (Dist.)} & \phantom{30}70.6 & (\texttt{\$502}) & \phantom{30}\textbf{58.9} & (\texttt{\$279}) & \phantom{30}74.2 & (\texttt{\$101}) & \phantom{0}80.9 & (\texttt{\$161}) & \phantom{30}\textbf{73.8} & (\texttt{\$245}) & \phantom{100}17.8 & (\texttt{\$148}) \\
\texttt{T5-Small (Ann.)} & 70.5 & (\texttt{\$1,300}) & \multicolumn{2}{c}{N/A} & 74.0 & (\texttt{\$1,032}) & 81.0 & (\texttt{\$1,980}) & \multicolumn{2}{c}{N/A} & 17.8 & (\texttt{\$3,321}) \\
\texttt{T5-Small (Ann.)} - Upper Bound & \textbf{71.1} &  (\texttt{\$1,800}) & 53.0 & (\texttt{\$2,595}) & \textbf{76.9} & (\texttt{\$12,899}) & \textbf{87.5} &  (\texttt{\$17,443}) & 69.3 & (\texttt{\$14,469}) & \textbf{26.2} & (\texttt{\$9,981}) \\
\bottomrule
\end{tabular}
\end{adjustbox}
\caption{Performances along with (the corresponding budget) of \texttt{Dist}., \texttt{Ann}. that performs the same/similar to \texttt{Dist}., and \texttt{Ann}. upper bound by leveraging all existing annotated data. The best performance for each task is in bold.}
\label{Table:analysis2_of_results}
\end{table*}

\subsection{\textbf{Annotation vs. Distillation}}
\label{main_results}
In Table \ref{Table:main}, we observe that interestingly, the \texttt{distillation (Dist.)} strategy significantly outperforms the annotation (\texttt{Ann}). strategy across almost all cases for all tasks. While knowledge distillation \cite{44873} has been proven effective for compression/generalization in previous works \cite{sanh2019distilbert, kang-etal-2020-regularization, Le2022FewShotAR}, our result that takes into account the realistic costs involved in building models is quite surprising, which highlights a new aspect: it is economically efficient. In other words, this suggests that exclusive reliance on scaling data by hiring human annotators might not be a good practice in light of cost efficiency.

Note that \texttt{Dist}. needs to be first fine-tuned on \colorbox{labeled}{\textit{N}} labeled data that requires a considerable computational cost, so if the fine-tuning cost exceeds the given budget, we denote such cases as N/A. In such scenarios, \texttt{Ann}. is essentially the right choice.
We also notice some scenarios where \texttt{Ann}. is a better option with limited budgets. For example, \texttt{Ann}. defeats its counterpart with \texttt{\$100} for WLP (\colorbox{labeled}{\textit{N}=\textit{5K}}) and \textsc{MultiPIT$_\texttt{Gen}$} (\colorbox{labeled}{\textit{N}=\textit{10K}}). In these cases, the \colorbox{unlabeled}{\textit{\#unlabeled data}} used for distillation are highly limited (\colorbox{unlabeled}{\textit{7K}} and \colorbox{unlabeled}{\textit{10K}}, respectively) as fine-tuning costs make up a substantial portion of limited budgets.

\subsection{\textbf{Does Distillation Work Better Simply by Making Use of Unlabeled Data?}}
In Table \ref{Table:main}, we observe a substantial performance gap between \texttt{Ann}. and \texttt{Dist}. One notable point is that there is a big difference in the absolute number of data (\colorbox{labeled}{\textit{\#labeled data}} and \colorbox{unlabeled}{\textit{\#unlabeled data}}) used for each strategy given a fixed budget. In Table \ref{Table:main}, for instance in \textsc{WLP}, given \texttt{\$500}, \colorbox{labeled}{\textit{1923}} more data can be annotated for \texttt{Ann}., whereas \colorbox{unlabeled}{\textit{111K}} unlabeled data can be leveraged for \texttt{Dist}. This not only means that annotated data is expensive, but also raises a question: \textit{is the performance gap simply because of the difference in the number of data points?} To investigate this question by building a fair ground in terms of the size of data, we take a \texttt{self-distillation (Self-Dist.)} approach \cite{zhang2019your} in which the architecture of a teacher and a student is the same (i.e., \texttt{T5-Small}).

In Table \ref{Table:analysis1_of_results}, we compare \texttt{Dist}. against \texttt{Self-Dist}. using the same \colorbox{unlabeled}{\textit{100K}} unlabeled data. We see that \texttt{Self-Dist}. is worse than the \texttt{Dist}. across all tasks by remarkable margins even though the same number of data is used. 
In fact, the performance of \texttt{Self-Dist}. is found to be bounded by its teacher (i.e., \texttt{T5-Small (Ann.)}), as also observed in \cite{Zhou2022PromptCF}.
This analysis suggests that the performance gap between \texttt{Dist}. and \texttt{Ann}. can indeed be attributed to exploiting the large pre-trained language model's capability, not simply making use of more data.

\subsection{Comparison under Larger Budgets}
Our experiments suggest that \texttt{distillation (Dist.)} is a more economical choice than relying completely on the human annotation to train a compact model, at least within scenarios presented in Table \ref{Table:main}.  However, this raises a question: \textit{could \texttt{Ann}. reach the performance of \texttt{Dist}. when investing a much larger budget?}  Table \ref{Table:analysis2_of_results} shows the results of \texttt{Dist}. with budgets for \colorbox{unlabeled}{\textit{100K}} unlabeled data, and \texttt{Ann}. with much larger budgets (or upper bound by using all available \colorbox{labeled}{\textit{\#labeled data}}). Interestingly, in some cases (\textsc{Stanceosaurus} \& \textsc{MultiPIT$_\texttt{Gen}$}), \texttt{Dist}. turns out to be an astoundingly economically efficient way to train a compact model. Even though all existing annotated data (\colorbox{labeled}{$\sim$\textit{50K}}) are used for \textsc{MultiPIT$_\texttt{Gen}$} training (w/ \texttt{\$14,469}), it never outperforms \texttt{Dist}. (w/ only \texttt{\$245}). For other tasks except for the aforementioned ones, we notice that \texttt{Ann}. can outperform \texttt{Dist}. with much larger budgets (e.g., \$\texttt{12,899} for FEVER). In practice, however, we still find that \texttt{Ann}. can be much more costly (e.g. 10x in the case of FEVER) to obtain similar performance.

\begin{figure*}[h!]
    \centering
    \includegraphics[scale=0.38]{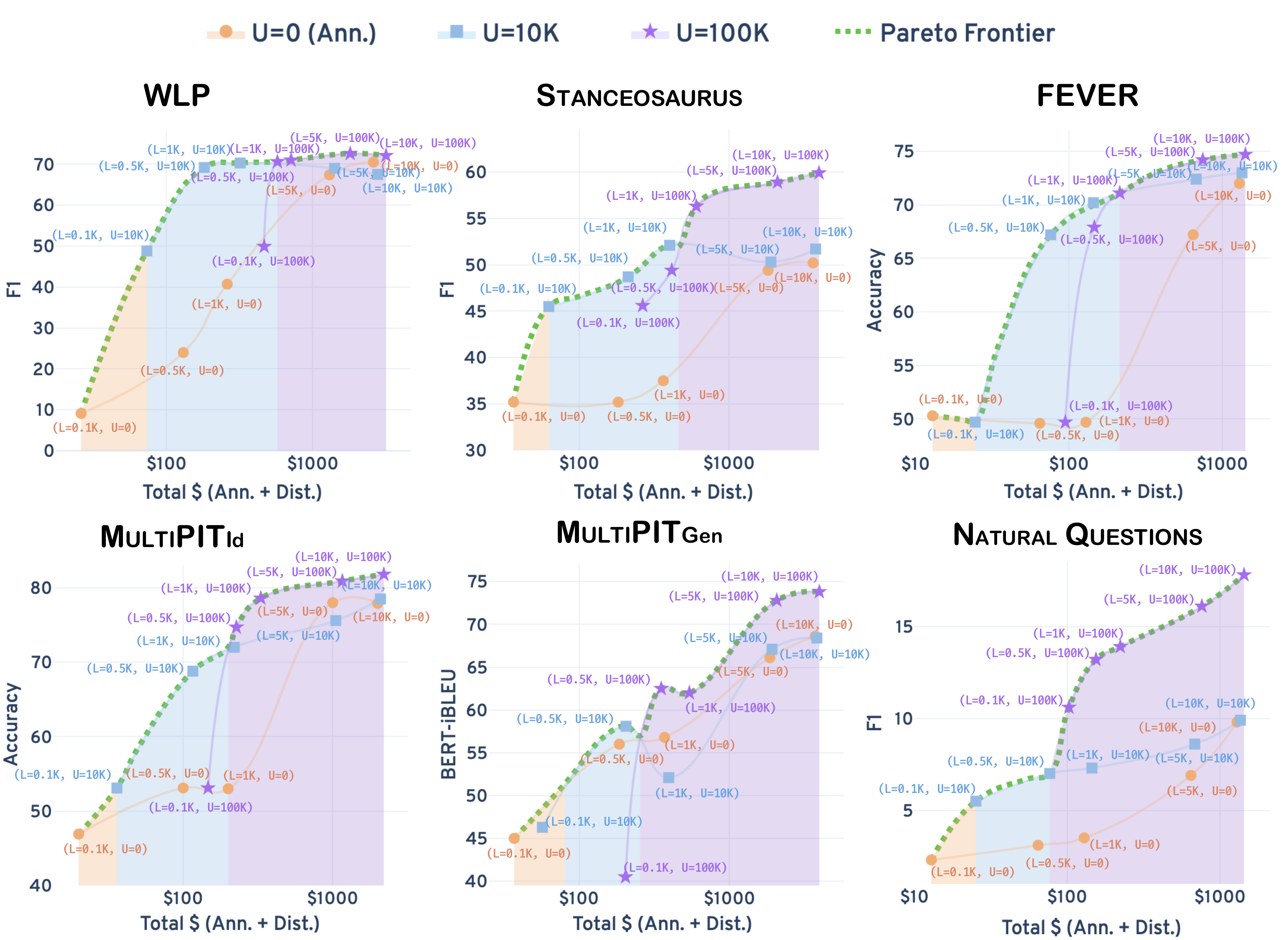}
    \caption{Pareto curves with various combinations of \colorbox{labeled}{\textit{\#labeled data}} (\texttt{L=\{0.1K}, \texttt{0.5K}, \texttt{1K}, \texttt{5K}, \texttt{10K\}}) and \colorbox{unlabeled}{\textit{\#unlabeled data}} (\texttt{U=\{\colorbox{u0}{0}}, \texttt{\colorbox{u10}{10K}}, \texttt{\colorbox{u100}{100K}\}}). \colorbox{u0}{\texttt{U=0}} denotes the \texttt{annotation (Ann.)} strategy. The Pareto frontier (\textcolor{kellygreen}{\small{$\sq\sq\sq\ \sq $}}) is the set of optimal solutions that practitioners would choose from, and is approximated by interpolating the given data points. 
    The X-axis is on a logarithmic scale.}
    \label{fig:pareto}
\end{figure*}

\section{Further Analyses}
In this section, we study varied values of each variable: the initial number (\textit{N}) of annotated data (\S \ref{sec:initial_data}), the compact model size (\S \ref{sec:compact_size}), and the teacher model size (\S \ref{sec:teacher_size}), all of which are fixed in the main experiment ({\S \ref{main_results}}).

\subsection{Pareto Curves}
\label{sec:initial_data}

In Figure \ref{fig:pareto}, we explore different combinations of \colorbox{labeled}{\textit{\#labeled data}} (\texttt{L=\{0.1K,} \texttt{0.5K}, \texttt{1K}, \texttt{5K}, \texttt{10K\}}) and \colorbox{unlabeled}{\textit{\#unlabeled data}} (\texttt{U=\{\colorbox{u0}{0},} \texttt{\colorbox{u10}{10K}}, \texttt{\colorbox{u100}{100K}\}}). Note that \colorbox{u0}{\texttt{U=0}} indicates the \texttt{annotation (Ann.)} strategy in essence. We plot the performances of each combination and approximate the Pareto frontier \citep{abdolrashidi2021pareto,treviso2022predicting} by interpolating the given data points.
For all tasks, we observe that the \texttt{distillation (Dist.)} strategy is almost always Pareto optimal.\footnote{One exception is \colorbox{u0}{\texttt{(L=0.1K, U=0)}} where a budget is so limited that leveraging a large model is barely feasible.}
In Appendix (Table \ref{Table:main_small_n}), we also look at the low resource setting in detail. 

Furthermore, we observe that using a smaller amount of unlabeled data (\colorbox{u10}{\texttt{U=10K}}) is Pareto optimal for smaller budgets, while larger unlabeled data (\colorbox{u100}{\texttt{U=100K}}) maximizes utility as the budget increases. This implies that in low-budget settings, the teacher's capacity is limited, allowing the student to catch up quickly. However, once the teacher outperforms the student by a significant margin, it is more economical to allocate a larger part of the budget towards distillation.

\begin{figure}[ht!]
    \centering
    \includegraphics[scale=0.23]{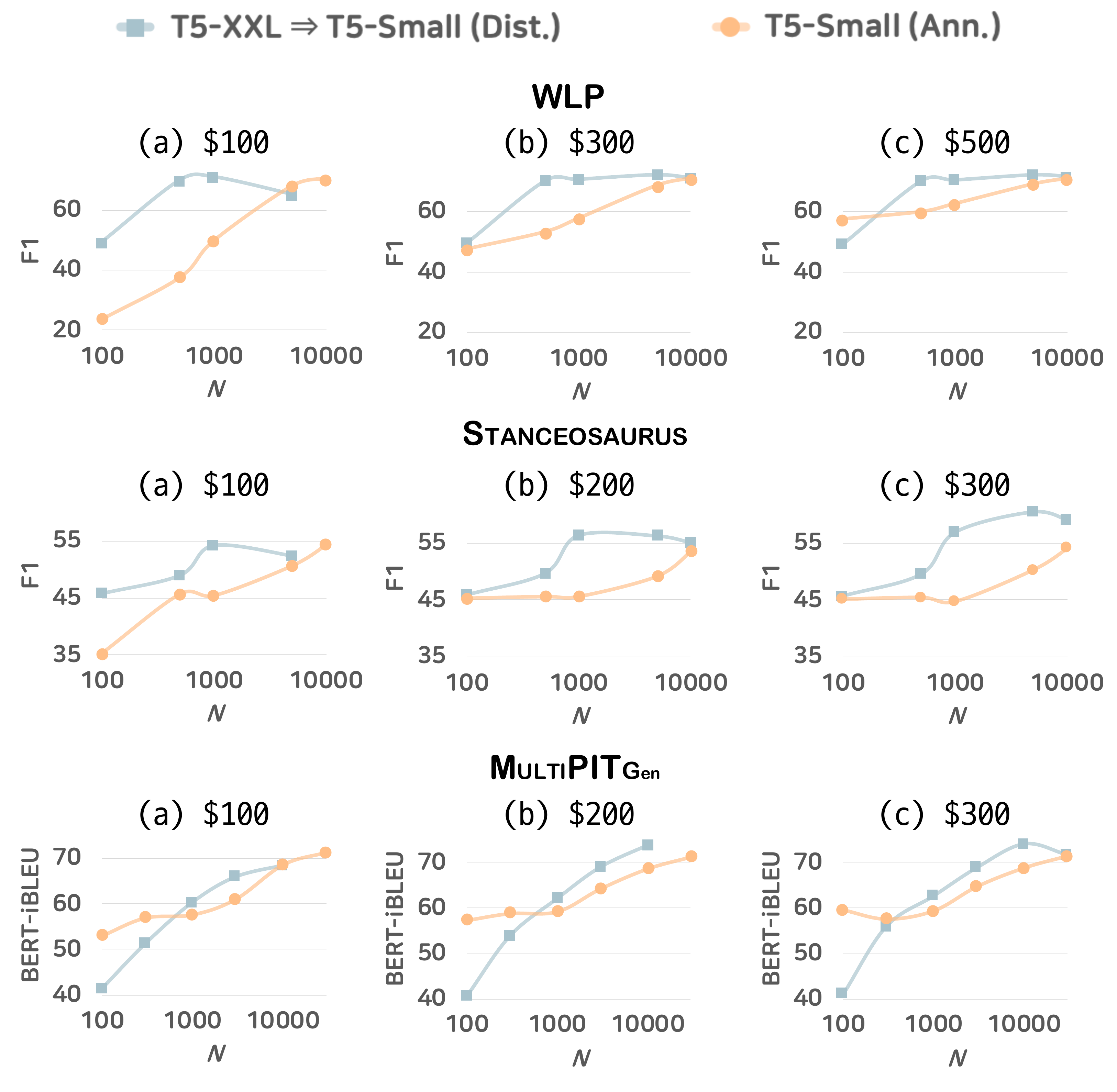}
    \caption{Results according to different number of starting annotated data ($N$) under fixed additional budgets.
    }
    \label{fig:diff_N}
\end{figure}

In Figure \ref{fig:diff_N}, we provide an additional analysis by varying the number of initially annotated data (\textit{N}) under fixed budgets to look at the impact of \textit{N}. Expectedly, we notice that \texttt{Dist}. outperforms \texttt{Ann}. in general except for some cases with low \textit{N}, especially for \textsc{MultiPIT$_\texttt{Gen}$} as also evidenced in Appendix (Table \ref{Table:main_small_n}). It is worth noting that there is a common trend across all tasks that the \texttt{Dist.} performances drop with high \textit{N}. This is due to the limited budgets; high \textit{N} requires a substantial fine-tuning cost for a large model, hence the budget to be used for distillation is limited. For instance, in the case of \textsc{Stanceosaurus} with \texttt{\textit{budget}=\$200}, if \textit{N} is \colorbox{labeled}{\textit{1K}}, \colorbox{unlabeled}{\textit{82K}} unlabeled data can be used for distillation, whereas only \colorbox{unlabeled}{\textit{35K}} unlabeled data are used when \textit{N}=\colorbox{labeled}{\textit{10K}}, resulting in the former outperforming the latter. This offers a lesson that unconditionally pursuing larger \textit{N} is not desirable in a fixed budget scenario; it is advisable for practitioners to understand and consider the trade-off between the fine-tuning and distillation costs.

\begin{figure}[t!]
    \centering
    \includegraphics[scale=0.31]{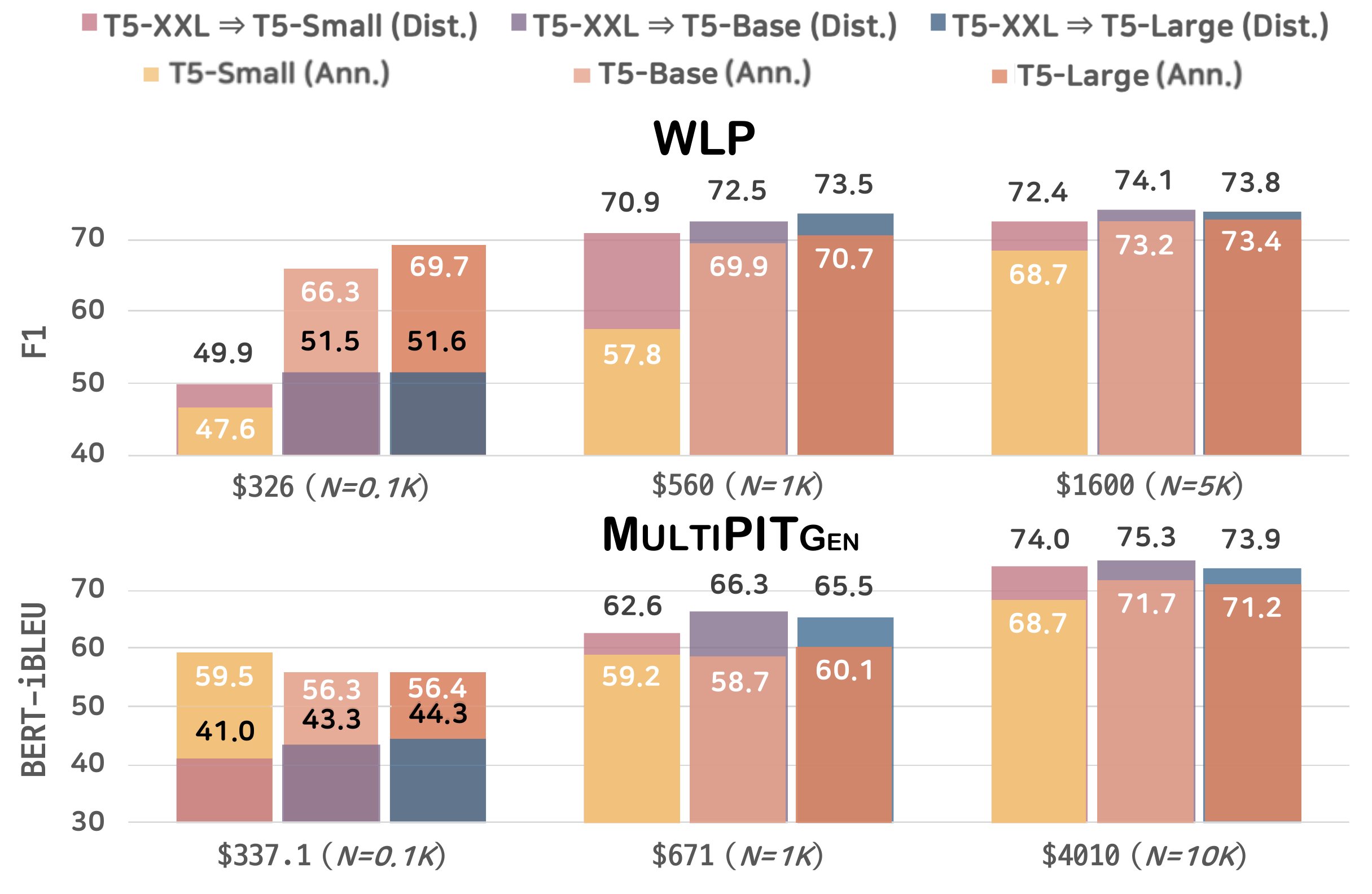}
    \caption{Results with different compact model sizes: \texttt{Small} (60M), \texttt{Base} (220M), \texttt{Large} (770M). For \texttt{Dist}., a teacher is fixed (\texttt{XXL-11B}), and the distillation cost is set to \texttt{\$300}. Best viewed in color.}
    \label{fig:diff_student}
\end{figure}

\subsection{\textbf{Varying the Compact Model Size}}
\label{sec:compact_size}
To consider various inference scenarios, we explore different sizes of a compact model in Figure \ref{fig:diff_student}. In general, the performances of all models improve as the budget increases, and \texttt{Dist}. outperforms \texttt{Ann}. given the same cost except for the low budget (\colorbox{labeled}{\textit{N=0.1K}}) setting. Interestingly, we observe that \texttt{T5-XXL} $\Rightarrow$ \texttt{T5-Base} \texttt{(Dist.)} is better than \texttt{T5-XXL} $\Rightarrow$ \texttt{T5-Large} \texttt{(Dist.)} in some cases (\texttt{\$1600} for WLP, \texttt{\$671} and \texttt{\$4010} for \textsc{MultiPIT}$_\texttt{Gen}$) although the former is smaller and more efficient. We conjecture that this is attributed to the model's larger number of parameters that require more GPUs and thereby more cost.
This result disproves the prevailing belief that larger models are always superior, at least in fixed-budget scenarios.

\begin{figure}[t!]
    \centering
    \includegraphics[scale=0.31]{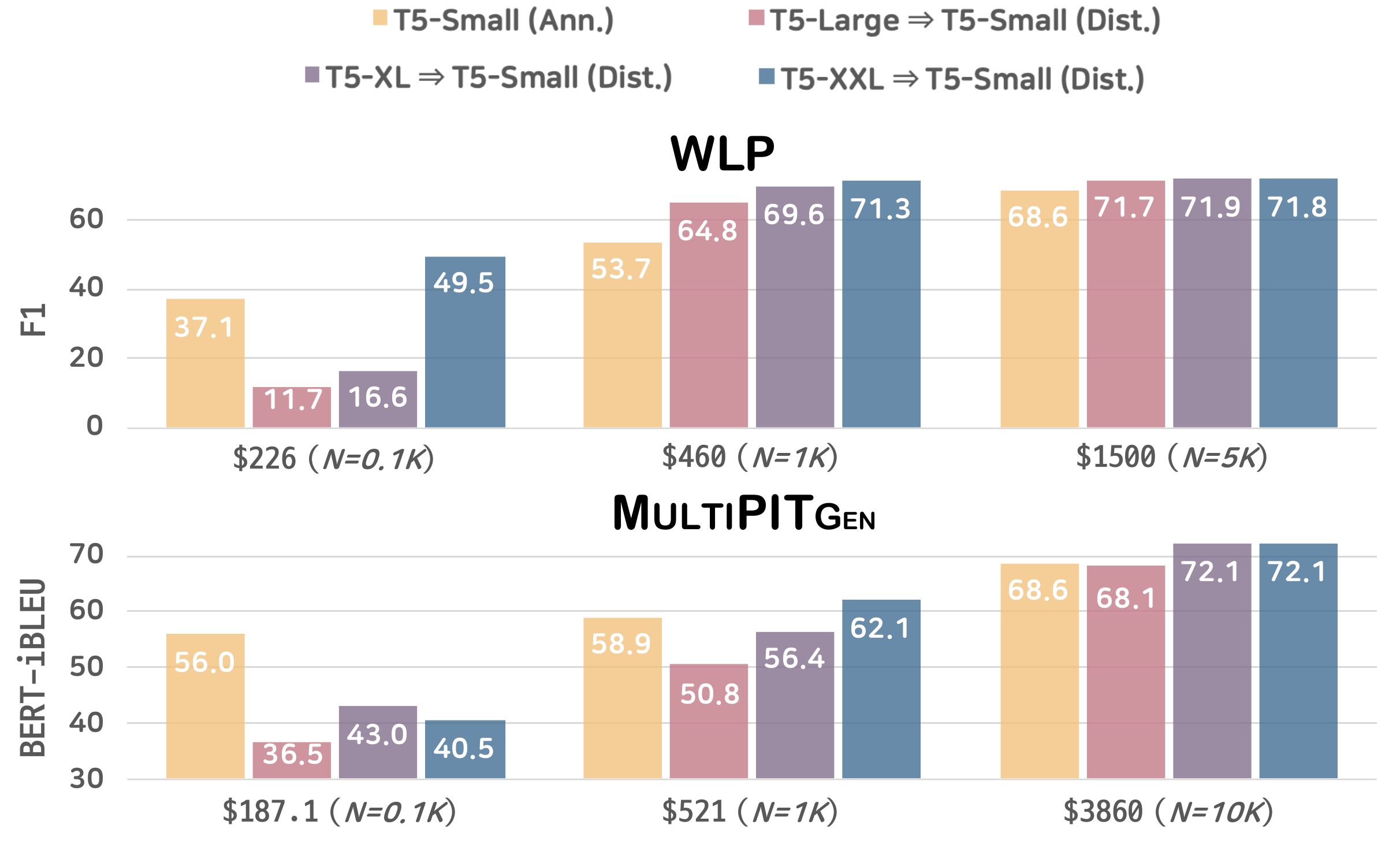}
    \caption{Results with varied scales of the teacher: \texttt{Large} (770M), \texttt{XL} (3B), \texttt{XXL} (11B). The compact model is fixed (\texttt{Small}-60M). The distillation cost is fixed as \texttt{\$200} for WLP and \texttt{\$150} for \textsc{MultiPIT}$_\texttt{Gen}$.}
    \label{fig:diff_teacher}
\end{figure}

\subsection{\textbf{Varying the Teacher Model Size}}
\label{sec:teacher_size}
We now investigate teacher models with different scales (Figure \ref{fig:diff_teacher}). It turns out that relatively smaller teacher models (\texttt{T5-Large} \& \texttt{T5-XL}) cannot be good teachers in the low budgets scenarios. For instance, with \texttt{\$521} for \textsc{MultiPIT}$_\texttt{Gen}$, \texttt{T5-Large} $\Rightarrow$ \texttt{T5-Small} \texttt{(Dist.)} and \texttt{T5-XL} $\Rightarrow$ \texttt{T5-Small} \texttt{(Dist.)} underperform \texttt{T5-Small (Ann.)}, whereas \texttt{T5-XXL} $\Rightarrow$ \texttt{T5-Small} \texttt{(Dist.)} outperforms \texttt{T5-Small (Ann.)}. In higher budget settings, it is noticeable that the largest teacher (\texttt{XXL}) is similar to or better than the smaller teacher (\texttt{Large}, \texttt{XL}). Taken together, this analysis suggests that when adopting distillation, the scale of the teacher model matters, and it may be safe to leverage sufficiently a larger model as a teacher regardless of any budgetary scenarios.

\section{\textbf{GPT-3.5 as an Annotator}}
Furthermore, we examine the cost efficiency of \texttt{GPT-3.5} \cite{ouyang2022training} annotation through an in-context few-shot learning scheme. \citet{wang-etal-2021-want-reduce} has recently demonstrated that \texttt{GPT-3} \cite{NEURIPS2020_1457c0d6} can be used as a cheaper labeler compared to humans. We attempt to scrutinize its applicability to the tasks considered in this work, and also contextualize its result with that of \texttt{Dist}. ultimately. We make use of the \texttt{text-davinci-003} model to generate pseudo-labels by prompting with 32 training examples. In this experiment, we assign \texttt{\$200} each for WLP and \textsc{Stanceosaurus} for \texttt{GPT-3.5} annotation. 
Note that OpenAI\footnote{https://openai.com/api/pricing} charges money based on the number of tokens used. The cost per label for WLP is \$0.046 and for \textsc{Stanceosaurus} is \$0.073, if using \texttt{GPT-3.5} (details in Appendix \ref{gpt-3.5_details}).

In Figure \ref{fig:gpt-3.5}, we compare \texttt{GPT-3.5} annotation (\texttt{GPT-3.5 Ann.}) against the human annotation and distillation strategy. In addition to \texttt{GPT-3.5 Ann.}, we combine it with human annotation (\texttt{Human + GPT-3.5 Ann.}) to enhance quality and make a comparison with \texttt{Dist}. The results clearly show that while \texttt{GPT-3.5} could be better than human annotators as hinted in prior work \cite{wang-etal-2021-want-reduce}, it significantly underperforms the distillation (\texttt{Dist.}) strategy given the same budget despite \texttt{GPT-3.5}'s larger parameters (175B) than the teacher (11B). This once again highlights the different view of knowledge distillation: cost efficiency.

\begin{figure}[t!]
    \centering
    \includegraphics[scale=0.38]{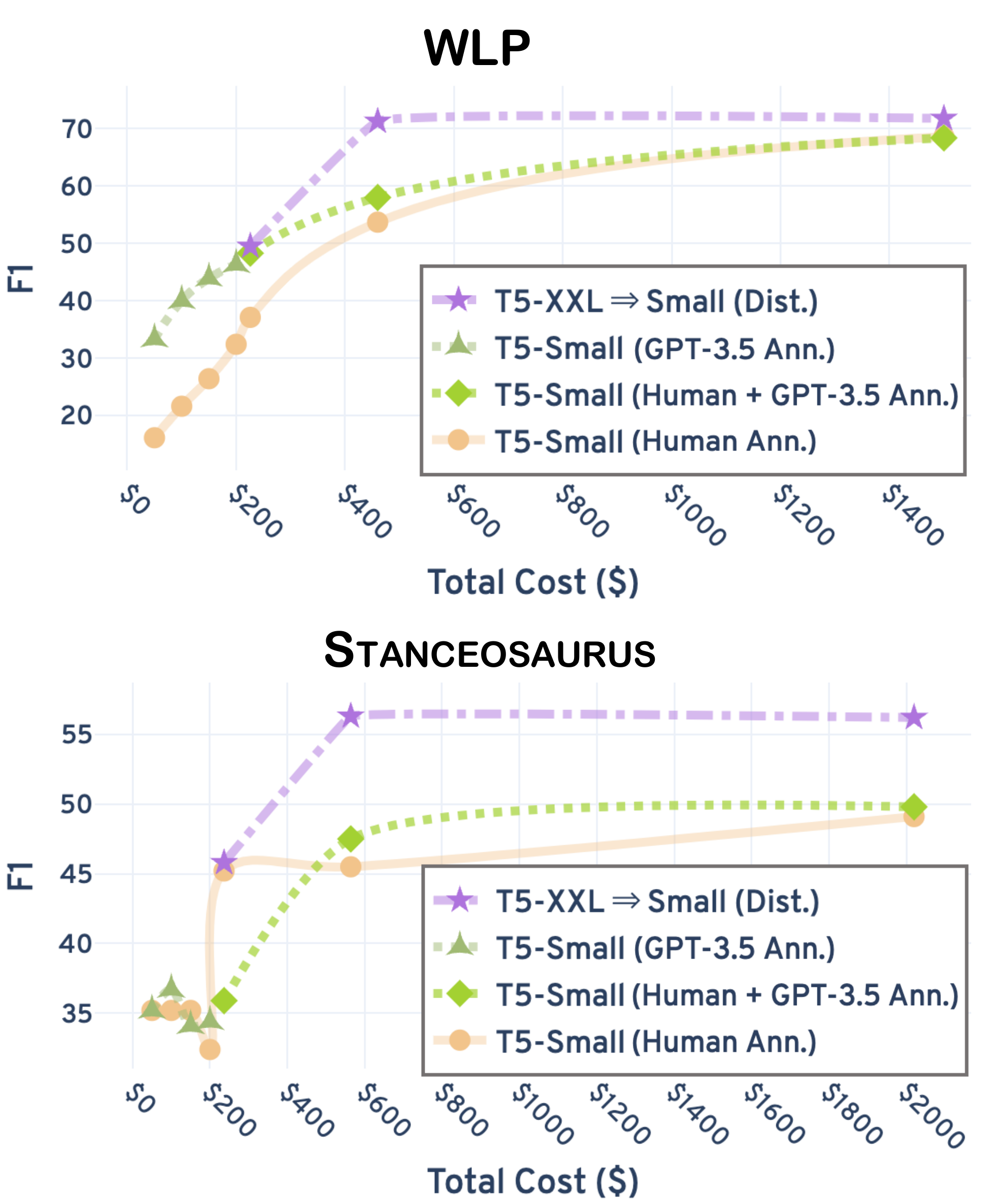}
    \caption{Comparisons with \texttt{GPT-3.5} annotation. Given an initial human annotation \textit{N}=\{\textit{0.1K}, \textit{1K}, \textit{5K}\} with the corresponding costs, \texttt{\$200} is additionally allocated for distillation or \texttt{GPT-3.5} annotation (i.e., \texttt{Human + GPT-3.5 Ann}.).}
    \label{fig:gpt-3.5}
\end{figure}

\section{Related Work}
The costs associated with building models have been explored or concerned by many prior works. 

\paragraph{Data Annotation.} On one hand, researchers have attempted to tackle the problem of noisy or expensive human annotation. For example, \citet{zhang2021learning} studies how to distribute annotation budgets between more examples with a single label and fewer examples with many labels. \citet{chen2022clean} investigates a redundant annotation with a majority vote vs. cleaning or relabeling the incorrect annotations. \citet{wang-etal-2021-want-reduce} compares human annotations against \texttt{GPT-3} \cite{NEURIPS2020_1457c0d6} annotations. However, these works only focus on the annotation cost.

\paragraph{Knowledge Distillation.} On the other hand, other lines of work address computational budgets associated with knowledge distillation. \citet{ye2022sparse} proposes using a larger and sparser student model than a teacher model to further reduce inference cost. \citet{jooste2022knowledge} compares different distillation schemes for cheap, fast, and environmentally friendly translation models. \citet{ma-etal-2022-knowledge} explores an efficient interactive distillation with meta-learning. The aforementioned works, however, ignore the data budgets and/or barely consider the realistic computational costs involved in the distillation process. While knowledge distillation has been shown effective for compression or generalization in previous NLP works \cite{sanh2019distilbert, kang-etal-2020-regularization, Le2022FewShotAR}, it remains unclear whether or not it is efficient even when considering the actual cost of distillation, which is often overlooked. As concurrent works, \citet{sun2023principledriven} presents a novel principle-driven self-alignment approach, and \citet{hsieh2023distilling} introduces a method that involves step-by-step distillation using chain-of-thought \cite{wei2022chain} rationales. Although the main focus is completely different from ours (i.e., cost), we believe that these works not only enhance this particular area but also have the potential to support our own findings regarding the cost-efficiency of distillation as the new methods would make the gap with annotation even bigger.

\paragraph{Data and Compute.} 

Unlike most existing works that consider exclusively either annotation or computational cost, our study contextualizes the two superficially dissociated types of costs, known to be expensive \cite{ning-etal-2019-partial, hong-etal-2020-handling, BENCHMARKS2021_6ea9ab1b, izsak-etal-2021-train, obando2020revisiting, minixhofer-etal-2022-wechsel} while being obscure in how they can be comparable to each other.
\citet{Kirstain2021AFM} compares scaling parameters against adding more labeled examples, but a compact model and a realistic cost (\$) are not of interest to it.
Our work resembles \citet{bai-etal-2021-pre} in terms of study framework, which explores how to optimally assign pre-training and annotation costs specifically for domain adaptation settings. Our focus is more on fine-tuning/distilling a compact model rather than pre-training from scratch and on exploring more general scenarios with diverse tasks.

\section{Conclusion}

In this work, we address a dilemma that practitioners often face when building a model: \textit{given a limited budget, how to invest it to train a compact model in an economically efficient manner?} 
We provide empirical evidence that (i) only scaling data using human annotators or \texttt{GPT-3.5} for annotation may not be the most economical solution, and (ii) when adopting the distillation strategy, using a smaller amount of unlabeled data leads to Pareto efficient models with a smaller budget, while it becomes more beneficial to use larger amounts of unlabeled data as the budget increases. Furthermore, (iii) we demonstrate that in budget-constrained settings, a smaller final model could produce both better performance and more efficient inference.
Given these findings, future work can explore different approaches to leveraging a large model's capability such as pruning for cost-efficient compact models.


\section*{Limitations}
This paper fundamentally considers a scenario in which practitioners rent cloud GPUs. In the case of hosting GPUs by themselves, the two strategies explored in this study would not be simply comparable. However, in practice, when training a large model (w/ 8 A100 GPUs), we conjecture that renting GPUs could be preferred in many cases as scaling compute powers is not trivial and prohibitively expensive \cite{izsak-etal-2021-train, obando2020revisiting, minixhofer-etal-2022-wechsel}. It is also noteworthy that in the future, computational costs may become cheaper as new hardware advances, the pricing policy by cloud platform services changes, and more optimization techniques are applied. On the other hand, human annotation cost is likely to be the same at least or even more expensive. With cost changes in such a direction, the same conclusion made by our study will hold even though the gap between the two strategies will get larger.

For a compression method, our work focuses on knowledge distillation \cite{44873}. However, it is worth noting that distillation amplifies a societal bias in a compressed model \cite{hooker2020characterising,silva-etal-2021-towards} due to its limited capacity \cite{ahn-etal-2022-knowledge}. Accordingly, practitioners are encouraged to additionally leverage bias mitigation techniques \cite{ahn-etal-2022-knowledge} when adopting distillation for real-world applications.
On top of our finding that the distillation scheme is more cost-efficient than the data annotation approach, other efficient methods such as pruning \cite{xia-etal-2022-structured} may be investigated in future work to decide which one is the best efficient solution among methods that leverages a large model. We believe, however, it should be noted that retaining performances after pruning a large portion (e.g., $\sim$99.995\%: 11B $\Rightarrow$ 60M) for a compact model would not be trivial, evidenced in a prior work \cite{NEURIPS2019_2c601ad9}.

\section*{Acknowledgments}
We thank Fan Bai and Jonathan Zheng for their assistance in estimating data annotation costs and collecting unlabeled data for WLP and Stanceosaurus, respectively.
This material is based upon work supported by the NSF (IIS-2052498) and by the Office of the Director of National Intelligence (ODNI), Intelligence Advanced Research Projects Activity (IARPA), via the HIATUS Program contract \#2022-22072200004. The views and conclusions contained herein are those of the authors and should not be interpreted as necessarily representing the official policies, either expressed or implied, of NSF, ODNI, IARPA, or the U.S. Government. The U.S. Government is authorized to reproduce and distribute reprints for governmental purposes notwithstanding any copyright annotation therein.

\bibliography{anthology,custom}

\begin{thebibliography}{63}
\expandafter\ifx\csname natexlab\endcsname\relax\def\natexlab#1{#1}\fi

\bibitem[{Abdolrashidi et~al.(2021)Abdolrashidi, Wang, Agrawal, Malmaud,
  Rybakov, Leichner, and Lew}]{abdolrashidi2021pareto}
AmirAli Abdolrashidi, Lisa Wang, Shivani Agrawal, Jonathan Malmaud, Oleg
  Rybakov, Chas Leichner, and Lukasz Lew. 2021.
\newblock Pareto-optimal quantized resnet is mostly 4-bit.
\newblock In \emph{Proceedings of the IEEE/CVF Conference on Computer Vision
  and Pattern Recognition}, pages 3091--3099.

\bibitem[{Ahn et~al.(2022)Ahn, Lee, Kim, and Oh}]{ahn-etal-2022-knowledge}
Jaimeen Ahn, Hwaran Lee, Jinhwa Kim, and Alice Oh. 2022.
\newblock \href {https://doi.org/10.18653/v1/2022.gebnlp-1.27} {Why knowledge
  distillation amplifies gender bias and how to mitigate from the perspective
  of {D}istil{BERT}}.
\newblock In \emph{Proceedings of the 4th Workshop on Gender Bias in Natural
  Language Processing (GeBNLP)}, pages 266--272, Seattle, Washington.
  Association for Computational Linguistics.

\bibitem[{Bai et~al.(2021)Bai, Ritter, and Xu}]{bai-etal-2021-pre}
Fan Bai, Alan Ritter, and Wei Xu. 2021.
\newblock \href {https://doi.org/10.18653/v1/2021.emnlp-main.409} {Pre-train or
  annotate? domain adaptation with a constrained budget}.
\newblock In \emph{Proceedings of the 2021 Conference on Empirical Methods in
  Natural Language Processing}, pages 5002--5015, Online and Punta Cana,
  Dominican Republic. Association for Computational Linguistics.

\bibitem[{Becker(1965)}]{10.2307/2228949}
Gary~S. Becker. 1965.
\newblock \href {http://www.jstor.org/stable/2228949} {A theory of the
  allocation of time}.
\newblock \emph{The Economic Journal}, 75(299):493--517.

\bibitem[{Brown et~al.(2020)Brown, Mann, Ryder, Subbiah, Kaplan, Dhariwal,
  Neelakantan, Shyam, Sastry, Askell, Agarwal, Herbert-Voss, Krueger, Henighan,
  Child, Ramesh, Ziegler, Wu, Winter, Hesse, Chen, Sigler, Litwin, Gray, Chess,
  Clark, Berner, McCandlish, Radford, Sutskever, and
  Amodei}]{NEURIPS2020_1457c0d6}
Tom Brown, Benjamin Mann, Nick Ryder, Melanie Subbiah, Jared~D Kaplan, Prafulla
  Dhariwal, Arvind Neelakantan, Pranav Shyam, Girish Sastry, Amanda Askell,
  Sandhini Agarwal, Ariel Herbert-Voss, Gretchen Krueger, Tom Henighan, Rewon
  Child, Aditya Ramesh, Daniel Ziegler, Jeffrey Wu, Clemens Winter, Chris
  Hesse, Mark Chen, Eric Sigler, Mateusz Litwin, Scott Gray, Benjamin Chess,
  Jack Clark, Christopher Berner, Sam McCandlish, Alec Radford, Ilya Sutskever,
  and Dario Amodei. 2020.
\newblock \href
  {https://proceedings.neurips.cc/paper/2020/file/1457c0d6bfcb4967418bfb8ac142f64a-Paper.pdf}
  {Language models are few-shot learners}.
\newblock In \emph{Advances in Neural Information Processing Systems},
  volume~33, pages 1877--1901. Curran Associates, Inc.

\bibitem[{Chen et~al.(2022)Chen, Yu, and Bowman}]{chen2022clean}
Derek Chen, Zhou Yu, and Samuel Bowman. 2022.
\newblock Clean or annotate: How to spend a limited data collection budget.
\newblock In \emph{Proceedings of the Third Workshop on Deep Learning for
  Low-Resource Natural Language Processing}, pages 152--168.

\bibitem[{Devlin et~al.(2019)Devlin, Chang, Lee, and
  Toutanova}]{devlin2019bert}
Jacob Devlin, Ming-Wei Chang, Kenton Lee, and Kristina Toutanova. 2019.
\newblock Bert: Pre-training of deep bidirectional transformers for language
  understanding.
\newblock In \emph{Proceedings of the 2019 Conference of the North American
  Chapter of the Association for Computational Linguistics: Human Language
  Technologies, Volume 1 (Long and Short Papers)}, pages 4171--4186.

\bibitem[{Dou et~al.(2022)Dou, Jiang, and Xu}]{Dou2022ImprovingLP}
Yao Dou, Chao Jiang, and Wei Xu. 2022.
\newblock Improving large-scale paraphrase acquisition and generation.
\newblock In \emph{In Proceedings of the 2022 Conference on Empirical Methods
  in Natural Language Processing}, Abu Dhabi, United Arab Emirates. Association
  for Computational Linguistics.

\bibitem[{Gou et~al.(2021)Gou, Yu, Maybank, and Tao}]{gou2021knowledge}
Jianping Gou, Baosheng Yu, Stephen~J Maybank, and Dacheng Tao. 2021.
\newblock Knowledge distillation: A survey.
\newblock \emph{International Journal of Computer Vision}, 129(6):1789--1819.

\bibitem[{Guu et~al.(2020)Guu, Lee, Tung, Pasupat, and
  Chang}]{pmlr-v119-guu20a}
Kelvin Guu, Kenton Lee, Zora Tung, Panupong Pasupat, and Mingwei Chang. 2020.
\newblock \href {https://proceedings.mlr.press/v119/guu20a.html} {Retrieval
  augmented language model pre-training}.
\newblock In \emph{Proceedings of the 37th International Conference on Machine
  Learning}, volume 119 of \emph{Proceedings of Machine Learning Research},
  pages 3929--3938. PMLR.

\bibitem[{Hendrycks et~al.(2021)Hendrycks, Burns, Chen, and
  Ball}]{BENCHMARKS2021_6ea9ab1b}
Dan Hendrycks, Collin Burns, Anya Chen, and Spencer Ball. 2021.
\newblock \href
  {https://datasets-benchmarks-proceedings.neurips.cc/paper/2021/file/6ea9ab1baa0efb9e19094440c317e21b-Paper-round1.pdf}
  {Cuad: An expert-annotated nlp dataset for legal contract review}.
\newblock In \emph{Proceedings of the Neural Information Processing Systems
  Track on Datasets and Benchmarks}, volume~1.

\bibitem[{Hernandez et~al.(2021)Hernandez, Kaplan, Henighan, and
  McCandlish}]{hernandez2021scaling}
Danny Hernandez, Jared Kaplan, Tom Henighan, and Sam McCandlish. 2021.
\newblock Scaling laws for transfer.
\newblock \emph{arXiv preprint arXiv:2102.01293}.

\bibitem[{Hinton et~al.(2015)Hinton, Vinyals, and Dean}]{44873}
Geoffrey Hinton, Oriol Vinyals, and Jeffrey Dean. 2015.
\newblock \href {http://arxiv.org/abs/1503.02531} {Distilling the knowledge in
  a neural network}.
\newblock In \emph{NIPS Deep Learning and Representation Learning Workshop}.

\bibitem[{Hong et~al.(2020)Hong, Kang, Lim, and
  Myaeng}]{hong-etal-2020-handling}
Giwon Hong, Junmo Kang, Doyeon Lim, and Sung-Hyon Myaeng. 2020.
\newblock \href {https://doi.org/10.18653/v1/2020.coling-main.306} {Handling
  anomalies of synthetic questions in unsupervised question answering}.
\newblock In \emph{Proceedings of the 28th International Conference on
  Computational Linguistics}, pages 3441--3448, Barcelona, Spain (Online).
  International Committee on Computational Linguistics.

\bibitem[{Hooker et~al.(2020)Hooker, Moorosi, Clark, Bengio, and
  Denton}]{hooker2020characterising}
Sara Hooker, Nyalleng Moorosi, Gregory Clark, Samy Bengio, and Emily Denton.
  2020.
\newblock \href {http://arxiv.org/abs/2010.03058} {Characterising bias in
  compressed models}.

\bibitem[{Hsieh et~al.(2023)Hsieh, Li, Yeh, Nakhost, Fujii, Ratner, Krishna,
  Lee, and Pfister}]{hsieh2023distilling}
Cheng-Yu Hsieh, Chun-Liang Li, Chih-Kuan Yeh, Hootan Nakhost, Yasuhisa Fujii,
  Alexander Ratner, Ranjay Krishna, Chen-Yu Lee, and Tomas Pfister. 2023.
\newblock \href {http://arxiv.org/abs/2305.02301} {Distilling step-by-step!
  outperforming larger language models with less training data and smaller
  model sizes}.

\bibitem[{Izsak et~al.(2021)Izsak, Berchansky, and
  Levy}]{izsak-etal-2021-train}
Peter Izsak, Moshe Berchansky, and Omer Levy. 2021.
\newblock \href {https://doi.org/10.18653/v1/2021.emnlp-main.831} {How to train
  {BERT} with an academic budget}.
\newblock In \emph{Proceedings of the 2021 Conference on Empirical Methods in
  Natural Language Processing}, pages 10644--10652, Online and Punta Cana,
  Dominican Republic. Association for Computational Linguistics.

\bibitem[{Jiao et~al.(2020)Jiao, Yin, Shang, Jiang, Chen, Li, Wang, and
  Liu}]{jiao2020tinybert}
Xiaoqi Jiao, Yichun Yin, Lifeng Shang, Xin Jiang, Xiao Chen, Linlin Li, Fang
  Wang, and Qun Liu. 2020.
\newblock Tinybert: Distilling bert for natural language understanding.
\newblock In \emph{Findings of the Association for Computational Linguistics:
  EMNLP 2020}, pages 4163--4174.

\bibitem[{Jooste et~al.(2022)Jooste, Way, Haque, and
  Superbo}]{jooste2022knowledge}
Wandri Jooste, Andy Way, Rejwanul Haque, and Riccardo Superbo. 2022.
\newblock Knowledge distillation for sustainable neural machine translation.
\newblock In \emph{Proceedings of the 15th Biennial Conference of the
  Association for Machine Translation in the Americas (Volume 2: Users and
  Providers Track and Government Track)}, pages 221--230.

\bibitem[{Kang et~al.(2020)Kang, Hong, Puerto San~Roman, and
  Myaeng}]{kang-etal-2020-regularization}
Junmo Kang, Giwon Hong, Haritz Puerto San~Roman, and Sung-Hyon Myaeng. 2020.
\newblock \href {https://doi.org/10.18653/v1/2020.findings-emnlp.293}
  {Regularization of distinct strategies for unsupervised question generation}.
\newblock In \emph{Findings of the Association for Computational Linguistics:
  EMNLP 2020}, pages 3266--3277, Online. Association for Computational
  Linguistics.

\bibitem[{Kirstain et~al.(2022)Kirstain, Lewis, Riedel, and
  Levy}]{Kirstain2021AFM}
Yuval Kirstain, Patrick Lewis, Sebastian Riedel, and Omer Levy. 2022.
\newblock A few more examples may be worth billions of parameters.
\newblock In \emph{Findings of the Association for Computational Linguistics:
  EMNLP 2022}, Abu Dhabi, United Arab Emirates. Association for Computational
  Linguistics.

\bibitem[{Kwiatkowski et~al.(2019)Kwiatkowski, Palomaki, Redfield, Collins,
  Parikh, Alberti, Epstein, Polosukhin, Devlin, Lee, Toutanova, Jones, Kelcey,
  Chang, Dai, Uszkoreit, Le, and Petrov}]{kwiatkowski-etal-2019-natural}
Tom Kwiatkowski, Jennimaria Palomaki, Olivia Redfield, Michael Collins, Ankur
  Parikh, Chris Alberti, Danielle Epstein, Illia Polosukhin, Jacob Devlin,
  Kenton Lee, Kristina Toutanova, Llion Jones, Matthew Kelcey, Ming-Wei Chang,
  Andrew~M. Dai, Jakob Uszkoreit, Quoc Le, and Slav Petrov. 2019.
\newblock \href {https://doi.org/10.1162/tacl_a_00276} {Natural questions: A
  benchmark for question answering research}.
\newblock \emph{Transactions of the Association for Computational Linguistics},
  7:452--466.

\bibitem[{Lancaster(1966)}]{10.2307/1828835}
Kelvin~J. Lancaster. 1966.
\newblock \href {http://www.jstor.org/stable/1828835} {A new approach to
  consumer theory}.
\newblock \emph{Journal of Political Economy}, 74(2):132--157.

\bibitem[{Le et~al.(2022)Le, Bai, and Ritter}]{Le2022FewShotAR}
Nghia Le, Fan Bai, and Alan Ritter. 2022.
\newblock Few-shot anaphora resolution in scientific protocols via mixtures of
  in-context experts.
\newblock In \emph{Findings of the Association for Computational Linguistics:
  EMNLP 2022}, Abu Dhabi, United Arab Emirates. Association for Computational
  Linguistics.

\bibitem[{Lee et~al.(2020)Lee, Li, Wang, Yih, Ma, and
  Khabsa}]{lee-etal-2020-language}
Nayeon Lee, Belinda~Z. Li, Sinong Wang, Wen-tau Yih, Hao Ma, and Madian Khabsa.
  2020.
\newblock \href {https://doi.org/10.18653/v1/2020.fever-1.5} {Language models
  as fact checkers?}
\newblock In \emph{Proceedings of the Third Workshop on Fact Extraction and
  VERification (FEVER)}, pages 36--41, Online. Association for Computational
  Linguistics.

\bibitem[{Lester et~al.(2021)Lester, Al-Rfou, and Constant}]{lester2021power}
Brian Lester, Rami Al-Rfou, and Noah Constant. 2021.
\newblock The power of scale for parameter-efficient prompt tuning.
\newblock In \emph{Proceedings of the 2021 Conference on Empirical Methods in
  Natural Language Processing}.

\bibitem[{Lewis et~al.(2019)Lewis, Liu, Goyal, Ghazvininejad, Mohamed, Levy,
  Stoyanov, and Zettlemoyer}]{lewis2019bart}
Mike Lewis, Yinhan Liu, Naman Goyal, Marjan Ghazvininejad, Abdelrahman Mohamed,
  Omer Levy, Ves Stoyanov, and Luke Zettlemoyer. 2019.
\newblock Bart: Denoising sequence-to-sequence pre-training for natural
  language generation, translation, and comprehension.
\newblock \emph{arXiv preprint arXiv:1910.13461}.

\bibitem[{Liu et~al.(2019)Liu, Ott, Goyal, Du, Joshi, Chen, Levy, Lewis,
  Zettlemoyer, and Stoyanov}]{liu2019roberta}
Yinhan Liu, Myle Ott, Naman Goyal, Jingfei Du, Mandar Joshi, Danqi Chen, Omer
  Levy, Mike Lewis, Luke Zettlemoyer, and Veselin Stoyanov. 2019.
\newblock Roberta: A robustly optimized bert pretraining approach.
\newblock \emph{arXiv preprint arXiv:1907.11692}.

\bibitem[{Ma et~al.(2022)Ma, Wang, Yu, and Zhang}]{ma-etal-2022-knowledge}
Xinge Ma, Jin Wang, Liang-Chih Yu, and Xuejie Zhang. 2022.
\newblock \href {https://aclanthology.org/2022.coling-1.435} {Knowledge
  distillation with reptile meta-learning for pretrained language model
  compression}.
\newblock In \emph{Proceedings of the 29th International Conference on
  Computational Linguistics}, pages 4907--4917, Gyeongju, Republic of Korea.
  International Committee on Computational Linguistics.

\bibitem[{Magister et~al.(2022)Magister, Mallinson, Adamek, Malmi, and
  Severyn}]{magister2022teaching}
Lucie~Charlotte Magister, Jonathan Mallinson, Jakub Adamek, Eric Malmi, and
  Aliaksei Severyn. 2022.
\newblock Teaching small language models to reason.
\newblock \emph{arXiv preprint arXiv:2212.08410}.

\bibitem[{Michel et~al.(2019)Michel, Levy, and Neubig}]{NEURIPS2019_2c601ad9}
Paul Michel, Omer Levy, and Graham Neubig. 2019.
\newblock \href
  {https://proceedings.neurips.cc/paper/2019/file/2c601ad9d2ff9bc8b282670cdd54f69f-Paper.pdf}
  {Are sixteen heads really better than one?}
\newblock In \emph{Advances in Neural Information Processing Systems},
  volume~32. Curran Associates, Inc.

\bibitem[{Min et~al.(2022)Min, Lewis, Hajishirzi, and
  Zettlemoyer}]{min-etal-2022-noisy}
Sewon Min, Mike Lewis, Hannaneh Hajishirzi, and Luke Zettlemoyer. 2022.
\newblock \href {https://doi.org/10.18653/v1/2022.acl-long.365} {Noisy channel
  language model prompting for few-shot text classification}.
\newblock In \emph{Proceedings of the 60th Annual Meeting of the Association
  for Computational Linguistics (Volume 1: Long Papers)}, pages 5316--5330,
  Dublin, Ireland. Association for Computational Linguistics.

\bibitem[{Minixhofer et~al.(2022)Minixhofer, Paischer, and
  Rekabsaz}]{minixhofer-etal-2022-wechsel}
Benjamin Minixhofer, Fabian Paischer, and Navid Rekabsaz. 2022.
\newblock \href {https://doi.org/10.18653/v1/2022.naacl-main.293} {{WECHSEL}:
  Effective initialization of subword embeddings for cross-lingual transfer of
  monolingual language models}.
\newblock In \emph{Proceedings of the 2022 Conference of the North American
  Chapter of the Association for Computational Linguistics: Human Language
  Technologies}, pages 3992--4006, Seattle, United States. Association for
  Computational Linguistics.

\bibitem[{Nguyen et~al.(2016)Nguyen, Rosenberg, Song, Gao, Tiwary, Majumder,
  and Deng}]{DBLP:conf/nips/NguyenRSGTMD16}
Tri Nguyen, Mir Rosenberg, Xia Song, Jianfeng Gao, Saurabh Tiwary, Rangan
  Majumder, and Li~Deng. 2016.
\newblock \href {http://ceur-ws.org/Vol-1773/CoCoNIPS\_2016\_paper9.pdf} {{MS}
  {MARCO:} {A} human generated machine reading comprehension dataset}.
\newblock In \emph{Proceedings of the Workshop on Cognitive Computation:
  Integrating neural and symbolic approaches 2016 co-located with the 30th
  Annual Conference on Neural Information Processing Systems {(NIPS} 2016),
  Barcelona, Spain, December 9, 2016}, volume 1773 of \emph{{CEUR} Workshop
  Proceedings}. CEUR-WS.org.

\bibitem[{Ning et~al.(2019)Ning, He, Fan, and Roth}]{ning-etal-2019-partial}
Qiang Ning, Hangfeng He, Chuchu Fan, and Dan Roth. 2019.
\newblock \href {https://doi.org/10.18653/v1/N19-1227} {Partial or complete,
  that{'}s the question}.
\newblock In \emph{Proceedings of the 2019 Conference of the North {A}merican
  Chapter of the Association for Computational Linguistics: Human Language
  Technologies, Volume 1 (Long and Short Papers)}, pages 2190--2200,
  Minneapolis, Minnesota. Association for Computational Linguistics.

\bibitem[{Niu et~al.(2021)Niu, Yavuz, Zhou, Keskar, Wang, and
  Xiong}]{niu-etal-2021-unsupervised}
Tong Niu, Semih Yavuz, Yingbo Zhou, Nitish~Shirish Keskar, Huan Wang, and
  Caiming Xiong. 2021.
\newblock \href {https://doi.org/10.18653/v1/2021.emnlp-main.417} {Unsupervised
  paraphrasing with pretrained language models}.
\newblock In \emph{Proceedings of the 2021 Conference on Empirical Methods in
  Natural Language Processing}, pages 5136--5150, Online and Punta Cana,
  Dominican Republic. Association for Computational Linguistics.

\bibitem[{Obando-Ceron and Castro(2021)}]{obando2020revisiting}
Johan~S Obando-Ceron and Pablo~Samuel Castro. 2021.
\newblock Revisiting rainbow: Promoting more insightful and inclusive deep
  reinforcement learning research.
\newblock In \emph{Proceedings of the 38th International Conference on Machine
  Learning}, Proceedings of Machine Learning Research. PMLR.

\bibitem[{Ouyang et~al.(2022)Ouyang, Wu, Jiang, Almeida, Wainwright, Mishkin,
  Zhang, Agarwal, Slama, Gray, Schulman, Hilton, Kelton, Miller, Simens,
  Askell, Welinder, Christiano, Leike, and Lowe}]{ouyang2022training}
Long Ouyang, Jeffrey Wu, Xu~Jiang, Diogo Almeida, Carroll Wainwright, Pamela
  Mishkin, Chong Zhang, Sandhini Agarwal, Katarina Slama, Alex Gray, John
  Schulman, Jacob Hilton, Fraser Kelton, Luke Miller, Maddie Simens, Amanda
  Askell, Peter Welinder, Paul Christiano, Jan Leike, and Ryan Lowe. 2022.
\newblock \href {https://openreview.net/forum?id=TG8KACxEON} {Training language
  models to follow instructions with human feedback}.
\newblock In \emph{Advances in Neural Information Processing Systems}.

\bibitem[{Patterson et~al.(2021)Patterson, Gonzalez, Le, Liang, Munguia,
  Rothchild, So, Texier, and Dean}]{patterson2021carbon}
David Patterson, Joseph Gonzalez, Quoc Le, Chen Liang, Lluis-Miquel Munguia,
  Daniel Rothchild, David So, Maud Texier, and Jeff Dean. 2021.
\newblock Carbon emissions and large neural network training.
\newblock \emph{arXiv preprint arXiv:2104.10350}.

\bibitem[{Petroni et~al.(2021)Petroni, Piktus, Fan, Lewis, Yazdani, De~Cao,
  Thorne, Jernite, Karpukhin, Maillard, Plachouras, Rockt{\"a}schel, and
  Riedel}]{petroni-etal-2021-kilt}
Fabio Petroni, Aleksandra Piktus, Angela Fan, Patrick Lewis, Majid Yazdani,
  Nicola De~Cao, James Thorne, Yacine Jernite, Vladimir Karpukhin, Jean
  Maillard, Vassilis Plachouras, Tim Rockt{\"a}schel, and Sebastian Riedel.
  2021.
\newblock \href {https://doi.org/10.18653/v1/2021.naacl-main.200} {{KILT}: a
  benchmark for knowledge intensive language tasks}.
\newblock In \emph{Proceedings of the 2021 Conference of the North American
  Chapter of the Association for Computational Linguistics: Human Language
  Technologies}, pages 2523--2544, Online. Association for Computational
  Linguistics.

\bibitem[{Raffel et~al.(2020)Raffel, Shazeer, Roberts, Lee, Narang, Matena,
  Zhou, Li, and Liu}]{2020t5}
Colin Raffel, Noam Shazeer, Adam Roberts, Katherine Lee, Sharan Narang, Michael
  Matena, Yanqi Zhou, Wei Li, and Peter~J. Liu. 2020.
\newblock \href {http://jmlr.org/papers/v21/20-074.html} {Exploring the limits
  of transfer learning with a unified text-to-text transformer}.
\newblock \emph{Journal of Machine Learning Research}, 21(140):1--67.

\bibitem[{Roberts et~al.(2020)Roberts, Raffel, and
  Shazeer}]{roberts-etal-2020-much}
Adam Roberts, Colin Raffel, and Noam Shazeer. 2020.
\newblock \href {https://doi.org/10.18653/v1/2020.emnlp-main.437} {How much
  knowledge can you pack into the parameters of a language model?}
\newblock In \emph{Proceedings of the 2020 Conference on Empirical Methods in
  Natural Language Processing (EMNLP)}, pages 5418--5426, Online. Association
  for Computational Linguistics.

\bibitem[{Sanh et~al.(2019)Sanh, Debut, Chaumond, and
  Wolf}]{sanh2019distilbert}
Victor Sanh, Lysandre Debut, Julien Chaumond, and Thomas Wolf. 2019.
\newblock Distilbert, a distilled version of bert: smaller, faster, cheaper and
  lighter.
\newblock In \emph{NeurIPS EMC\^2 Workshop}.

\bibitem[{Schuster et~al.(2021)Schuster, Fisch, and
  Barzilay}]{schuster-etal-2021-get}
Tal Schuster, Adam Fisch, and Regina Barzilay. 2021.
\newblock \href {https://doi.org/10.18653/v1/2021.naacl-main.52} {Get your
  vitamin {C}! robust fact verification with contrastive evidence}.
\newblock In \emph{Proceedings of the 2021 Conference of the North American
  Chapter of the Association for Computational Linguistics: Human Language
  Technologies}, pages 624--643, Online. Association for Computational
  Linguistics.

\bibitem[{Silva et~al.(2021)Silva, Tambwekar, and
  Gombolay}]{silva-etal-2021-towards}
Andrew Silva, Pradyumna Tambwekar, and Matthew Gombolay. 2021.
\newblock \href {https://doi.org/10.18653/v1/2021.naacl-main.189} {Towards a
  comprehensive understanding and accurate evaluation of societal biases in
  pre-trained transformers}.
\newblock In \emph{Proceedings of the 2021 Conference of the North American
  Chapter of the Association for Computational Linguistics: Human Language
  Technologies}, pages 2383--2389, Online. Association for Computational
  Linguistics.

\bibitem[{Stanton et~al.(2021)Stanton, Izmailov, Kirichenko, Alemi, and
  Wilson}]{stanton2021does}
Samuel Stanton, Pavel Izmailov, Polina Kirichenko, Alexander~A Alemi, and
  Andrew~G Wilson. 2021.
\newblock Does knowledge distillation really work?
\newblock \emph{Advances in Neural Information Processing Systems}, 34.

\bibitem[{Sun et~al.(2023)Sun, Shen, Zhou, Zhang, Chen, Cox, Yang, and
  Gan}]{sun2023principledriven}
Zhiqing Sun, Yikang Shen, Qinhong Zhou, Hongxin Zhang, Zhenfang Chen, David
  Cox, Yiming Yang, and Chuang Gan. 2023.
\newblock \href {http://arxiv.org/abs/2305.03047} {Principle-driven
  self-alignment of language models from scratch with minimal human
  supervision}.

\bibitem[{Tabassum et~al.(2020)Tabassum, Xu, and
  Ritter}]{tabassum-etal-2020-wnut}
Jeniya Tabassum, Wei Xu, and Alan Ritter. 2020.
\newblock \href {https://doi.org/10.18653/v1/2020.wnut-1.33} {{WNUT}-2020 task
  1 overview: Extracting entities and relations from wet lab protocols}.
\newblock In \emph{Proceedings of the Sixth Workshop on Noisy User-generated
  Text (W-NUT 2020)}, pages 260--267, Online. Association for Computational
  Linguistics.

\bibitem[{Thorne et~al.(2018)Thorne, Vlachos, Christodoulopoulos, and
  Mittal}]{thorne-etal-2018-fever}
James Thorne, Andreas Vlachos, Christos Christodoulopoulos, and Arpit Mittal.
  2018.
\newblock \href {https://doi.org/10.18653/v1/N18-1074} {{FEVER}: a large-scale
  dataset for fact extraction and {VER}ification}.
\newblock In \emph{Proceedings of the 2018 Conference of the North {A}merican
  Chapter of the Association for Computational Linguistics: Human Language
  Technologies, Volume 1 (Long Papers)}, pages 809--819, New Orleans,
  Louisiana. Association for Computational Linguistics.

\bibitem[{Treviso et~al.(2022)Treviso, G{\'o}is, Fernandes, Fonseca, and
  Martins}]{treviso2022predicting}
Marcos Treviso, Ant{\'o}nio G{\'o}is, Patrick Fernandes, Erick Fonseca, and
  Andr{\'e}~FT Martins. 2022.
\newblock Predicting attention sparsity in transformers.
\newblock In \emph{Proceedings of the Sixth Workshop on Structured Prediction
  for NLP}, pages 67--81.

\bibitem[{Wang et~al.(2021)Wang, Liu, Xu, Zhu, and
  Zeng}]{wang-etal-2021-want-reduce}
Shuohang Wang, Yang Liu, Yichong Xu, Chenguang Zhu, and Michael Zeng. 2021.
\newblock \href {https://doi.org/10.18653/v1/2021.findings-emnlp.354} {Want to
  reduce labeling cost? {GPT}-3 can help}.
\newblock In \emph{Findings of the Association for Computational Linguistics:
  EMNLP 2021}, pages 4195--4205, Punta Cana, Dominican Republic. Association
  for Computational Linguistics.

\bibitem[{Wei et~al.(2022)Wei, Wang, Schuurmans, Bosma, brian ichter, Xia, Chi,
  Le, and Zhou}]{wei2022chain}
Jason Wei, Xuezhi Wang, Dale Schuurmans, Maarten Bosma, brian ichter, Fei Xia,
  Ed~H. Chi, Quoc~V Le, and Denny Zhou. 2022.
\newblock \href {https://openreview.net/forum?id=_VjQlMeSB_J} {Chain of thought
  prompting elicits reasoning in large language models}.
\newblock In \emph{Advances in Neural Information Processing Systems}.

\bibitem[{Wolf et~al.(2019)Wolf, Debut, Sanh, Chaumond, Delangue, Moi, Cistac,
  Rault, Louf, Funtowicz, Davison, Shleifer, von Platen, Ma, Jernite, Plu, Xu,
  Scao, Gugger, Drame, Lhoest, and Rush}]{wolf2019huggingfaces}
Thomas Wolf, Lysandre Debut, Victor Sanh, Julien Chaumond, Clement Delangue,
  Anthony Moi, Pierric Cistac, Tim Rault, Rémi Louf, Morgan Funtowicz, Joe
  Davison, Sam Shleifer, Patrick von Platen, Clara Ma, Yacine Jernite, Julien
  Plu, Canwen Xu, Teven~Le Scao, Sylvain Gugger, Mariama Drame, Quentin Lhoest,
  and Alexander~M. Rush. 2019.
\newblock \href {http://arxiv.org/abs/1910.03771} {Huggingface's transformers:
  State-of-the-art natural language processing}.

\bibitem[{Xia et~al.(2022)Xia, Zhong, and Chen}]{xia-etal-2022-structured}
Mengzhou Xia, Zexuan Zhong, and Danqi Chen. 2022.
\newblock \href {https://doi.org/10.18653/v1/2022.acl-long.107} {Structured
  pruning learns compact and accurate models}.
\newblock In \emph{Proceedings of the 60th Annual Meeting of the Association
  for Computational Linguistics (Volume 1: Long Papers)}, pages 1513--1528,
  Dublin, Ireland. Association for Computational Linguistics.

\bibitem[{Xu and McAuley(2022)}]{xu2022survey}
Canwen Xu and Julian McAuley. 2022.
\newblock \href {http://arxiv.org/abs/2202.07105} {A survey on model
  compression and acceleration for pretrained language models}.

\bibitem[{Ye et~al.(2022)Ye, Khabsa, Lewis, Wang, Ren, and
  Jaech}]{ye2022sparse}
Qinyuan Ye, Madian Khabsa, Mike Lewis, Sinong Wang, Xiang Ren, and Aaron Jaech.
  2022.
\newblock Sparse distillation: Speeding up text classification by using bigger
  student models.
\newblock In \emph{Proceedings of the 2022 Conference of the North American
  Chapter of the Association for Computational Linguistics: Human Language
  Technologies}, pages 2361--2375.

\bibitem[{Zhang et~al.(2019)Zhang, Song, Gao, Chen, Bao, and
  Ma}]{zhang2019your}
Linfeng Zhang, Jiebo Song, Anni Gao, Jingwei Chen, Chenglong Bao, and Kaisheng
  Ma. 2019.
\newblock Be your own teacher: Improve the performance of convolutional neural
  networks via self distillation.
\newblock In \emph{Proceedings of the IEEE/CVF International Conference on
  Computer Vision}, pages 3713--3722.

\bibitem[{Zhang et~al.(2021)Zhang, Gong, and Choi}]{zhang2021learning}
Shujian Zhang, Chengyue Gong, and Eunsol Choi. 2021.
\newblock Learning with different amounts of annotation: From zero to many
  labels.
\newblock In \emph{Proceedings of the 2021 Conference on Empirical Methods in
  Natural Language Processing}, pages 7620--7632.

\bibitem[{Zhang et~al.(2020)Zhang, Kishore, Wu, Weinberger, and
  Artzi}]{Zhang*2020BERTScore:}
Tianyi Zhang, Varsha Kishore, Felix Wu, Kilian~Q. Weinberger, and Yoav Artzi.
  2020.
\newblock \href {https://openreview.net/forum?id=SkeHuCVFDr} {Bertscore:
  Evaluating text generation with bert}.
\newblock In \emph{International Conference on Learning Representations}.

\bibitem[{Zhao et~al.(2021)Zhao, Wallace, Feng, Klein, and
  Singh}]{pmlr-v139-zhao21c}
Zihao Zhao, Eric Wallace, Shi Feng, Dan Klein, and Sameer Singh. 2021.
\newblock \href {https://proceedings.mlr.press/v139/zhao21c.html} {Calibrate
  before use: Improving few-shot performance of language models}.
\newblock In \emph{Proceedings of the 38th International Conference on Machine
  Learning}, volume 139 of \emph{Proceedings of Machine Learning Research},
  pages 12697--12706. PMLR.

\bibitem[{Zheng et~al.(2022)Zheng, Baheti, Naous, Xu, and
  Ritter}]{Zheng2022StanceosaurusCS}
Jonathan Zheng, Ashutosh Baheti, Tarek Naous, Wei Xu, and Alan Ritter. 2022.
\newblock Stanceosaurus: Classifying stance towards multilingual
  misinformation.
\newblock In \emph{In Proceedings of the 2022 Conference on Empirical Methods
  in Natural Language Processing}, Abu Dhabi, United Arab Emirates. Association
  for Computational Linguistics.

\bibitem[{Zhou et~al.(2022{\natexlab{a}})Zhou, He, Ma, Berg-Kirkpatrick, and
  Neubig}]{Zhou2022PromptCF}
Chunting Zhou, Junxian He, Xuezhe Ma, Taylor Berg-Kirkpatrick, and Graham
  Neubig. 2022{\natexlab{a}}.
\newblock Prompt consistency for zero-shot task generalization.

\bibitem[{Zhou et~al.(2022{\natexlab{b}})Zhou, Xu, and McAuley}]{zhou2022bert}
Wangchunshu Zhou, Canwen Xu, and Julian McAuley. 2022{\natexlab{b}}.
\newblock Bert learns to teach: Knowledge distillation with meta learning.
\newblock In \emph{Proceedings of the 60th Annual Meeting of the Association
  for Computational Linguistics (Volume 1: Long Papers)}, pages 7037--7049.

\end{thebibliography}
\bibliographystyle{acl_natbib}
\clearpage
\appendix

\section{Details of Annotation Cost Estimation}
\label{ann-cost-details}
\paragraph{WLP {\normalfont \cite{tabassum-etal-2020-wnut}}} This is an annotated corpus containing wet lab protocols, and the included tasks are named entity recognition (NER) and relation extraction (RE). We refer to \citet{bai-etal-2021-pre} for the price per sentence (instance), which is \$0.44. Since this price is measured for both tasks, and we are only interested in NER, we take the ratio of the number of labels for each (59.76\%:40.24\%) for the estimate of NER in isolation, yielding approximately \$0.26.
\paragraph{\textsc{Stanceosaurus} {\normalfont \cite{Zheng2022StanceosaurusCS}}} This dataset includes sourced claims and relevant tweets along with annotated stances for stance classification. Since the labeling cost was not explicitly mentioned in the paper, we asked the authors for the details of the average number of annotations per hour (82 tweets) and the hiring cost (\$15 per hour) to calculate the final price per label: \$15 $\div$ 82 $\times$ 2 (double-annotated) = \$0.364.
\paragraph{\textsc{MultiPIT} {\normalfont \cite{Dou2022ImprovingLP}}} This provides Twitter-based paraphrase containing multiple topics. We specifically consider, out of variants, \textsc{MultiPIT}\textsubscript{CROWD} corpus, consisting of sentence pairs labeled whether each pair is paraphrased or not for paraphrase identification (\textsc{MultiPIT$_\texttt{Id}$}). The cost per pair is considered \$0.2 as mentioned in the paper. For paraphrase generation (\textsc{MultiPIT$_\texttt{Gen}$}), we sample pairs annotated as paraphrased, and take the proportion of sampled ones out of the total (53.9\%) to get the cost per paraphrased source-target instance: 100 $\div$ 53.9 $\times$ \$0.2 = \$0.371.
\paragraph{FEVER {\normalfont \cite{thorne-etal-2018-fever}} \& \textsc{Natural Questions} {\normalfont \cite{kwiatkowski-etal-2019-natural}}} These are fact verification and question answering datasets respectively for which we estimate the costs by leveraging the price from Google Cloud Platform. This charges \$129 per 50 words for 1,000 units, and hence we get an estimate of \$0.129 per label for both tasks.

\section{Input-Output Formats for Each Task}
\label{input-output}
Our study uses T5 \cite{2020t5} as our base model under the standard text-to-text framework. The input-output examples for each task are demonstrated in Table \ref{Table:input-output}, and what follows is detailed explanations for each.
\paragraph{WLP} This task can be regarded as a token-level classification problem, where the \#class is 20 in total: \{Amount, Reagent, Device, Time, Speed, Action, Mention, Location, Numerical, Method, Temperature, Modifier, Concentration, Size, Generic-Measure, Seal, Measure-Type, Misc, Ph, Unit\}. Given a source input (i.e., procedural sentence), the model is required to generate a target as a form of "Entity [Label] Entity [Label] ...". 

\paragraph{\textsc{Stanceosaurus}} For this task, the source is the concatenation of a claim, a relevant tweet, and context information (e.g., reply), and the target is supposed to one of \{Supporting | Refuting | Irrelevant | Discussing | Querying\}. 

\paragraph{FEVER} This is a fact verification task where the source is a claim (closed-book setting as discussed in \S \ref{settings}), and the target is Supports or Refutes in a 2-way classification setting following \citet{petroni-etal-2021-kilt}. 

\paragraph{\textsc{MultiPIT$_\texttt{Id}$}} is also a binary classification task where given two sentences, targets should be Yes or No.

\paragraph{\textsc{MultiPIT$_\texttt{Gen}$}} The source for this task is a sentence and the target is a paraphrased sentence. 

\paragraph{\textsc{Natural Questions}} As in \textbf{FEVER}, we also consider the closed-book setup that requires a model to rely on its implicit knowledge for this task where the question is a source and the target is directly the answer to the question.

\begin{table}[h!]
\centering
\begin{adjustbox}{width=0.48\textwidth}
\begin{tabular}{lcccc}
\toprule
\textbf{Dataset} & \textbf{\#Train} & \textbf{\#Dev} & \textbf{\#Test} & {\textbf{\#Unlabeled Data}} \\
\midrule\midrule
\textbf{WLP} & 11,966 & 2,861 & 3,562 & 111,000 \\
\textbf{\textsc{Stanceosaurus}} & 12,130 & 3,827 & 4,750 & 126,000 \\
\textbf{FEVER} & 104,966 & 10,444 & N/A & 182,000 \\
\textbf{\textsc{MultiPIT$_\texttt{Id}$}} & 92,217 & 11,527 & 11,530 & 237,000 \\
\textbf{\textsc{MultiPIT$_\texttt{Gen}$}} & 49,673 & 6,143 & 6,120 & 179,000 \\
\textbf{\textsc{Natural Questions}} & 87,372 & 2,837 &	N/A & 115,000 \\
\bottomrule
\end{tabular}
\end{adjustbox}
\caption{Statistics for various NLP datasets. For FEVER and \textsc{Natural Questions}, dev sets are used for evaluation as test sets are private. The maximum number of unlabeled data used for experiments is presented.
}
\label{Table:data_statistics}
\end{table}

\begin{table*}[th!]
\centering
\begin{adjustbox}{width=0.7\textwidth}
\begin{tabular}{lccccc}
\toprule
\multirow{2}{*}{\textbf{Dataset}} & \textbf{Initial \$} & \textbf{\texttt{Annotation (Ann.)}} & \textbf{\texttt{Distillation (Dist.)}} \\
 & \footnotesize{(\colorbox{labeled}{\textit{N=5K}})} & \footnotesize{\textbf{\texttt{T5-Small}} (\colorbox{labeled}{\textit{+1K}})} & \footnotesize{\textbf{\texttt{T5-XXL}} (\colorbox{labeled}{\textit{5K}}) $\Rightarrow$ \textbf{\texttt{T5-Small}} (\colorbox{unlabeled}{\textit{100K}})} \\
\midrule\midrule
\textbf{WLP} & \texttt{\$1,300} & \texttt{\$260} & \texttt{\$67.5 $\Rightarrow$ \$435} \\
\textbf{\textsc{Stanceosaurus}} & \texttt{\$1,820} & \texttt{\$364} & \texttt{\$60 $\Rightarrow$ \$225} \\
\textbf{FEVER} & \texttt{\$645} & \texttt{\$129} & \texttt{\$22.5 $\Rightarrow$ \$78} \\
\textbf{\textsc{MultiPIT$_\texttt{Id}$}} & \texttt{\$1,000} & \texttt{\$200} & \texttt{\$37.5 $\Rightarrow$ \$123} \\
\textbf{\textsc{MultiPIT$_\texttt{Gen}$}} & \texttt{\$1,855} & \texttt{\$371} & \texttt{\$45 $\Rightarrow$ \$163} \\
\textbf{\textsc{Natural Questions}} & \texttt{\$645} & \texttt{\$129} & \texttt{\$30 $\Rightarrow$ \$86} \\
\bottomrule
\end{tabular}
\end{adjustbox}
\caption{Example breakdown of cost estimates for training compact models using the two approaches illustrated in Figure \ref{fig:strategies}.
Starting with (\colorbox{labeled}{\textit{\footnotesize{5K}}}) labeled examples, we compare the costs of annotating an additional \colorbox{labeled}{\textit{\footnotesize{+1K}}}, or fine-tuning, then distilling \texttt{T5-XXL} (11B parameters).  For \textbf{\texttt{\footnotesize{Distillation (Dist.)}}}, the computational cost involves fine-tuning \texttt{T5-XXL} (the teacher) on \colorbox{labeled}{\textit{\footnotesize{5K}}} annotated data, plus distilling it into \texttt{T5-Small} using \colorbox{unlabeled}{\textit{\footnotesize{100K}}} unlabeled examples.
}
\label{Table:compute_cost_analysis}
\end{table*}

\begin{table*}[ht!]
\centering
\begin{adjustbox}{width=1\textwidth}
\begin{tabular}{lll}
\toprule
\textbf{Dataset} & \textbf{Task} & \textbf{Example} \\
\midrule\midrule
\multirow{2}{*}{\textbf{WLP}}  & \multirow{2}{*}{\footnotesize{Named Entity Recognition}} & \textbf{Source - } \footnotesize{Assemble the following reagents in a thin-walled PCR tube} \\
\cmidrule(lr){3-3} & & \textbf{Target - } \footnotesize{Assemble [Action] following reagents [Reagent] thin-walled PCR tube [Location]} \\
\midrule

\multirow{5.5}{*}{\textbf{\textsc{Stanceosaurus}}}  & \multirow{5.5}{*}{\footnotesize{Stance Classification}} & \textbf{Source - } \footnotesize{claim: The suicide rate increased during COVID-19 lockdown.} \\ 
& & \footnotesize{\quad\quad\quad\quad\enspace\, tweet: @USER @USER People who are suicidal can hide the signs very well.} \\
& & \footnotesize{\quad\quad\quad\quad\enspace\, [SEP] @USER @USER So we aren't looking at the family units for this then? If people are at home all day, } \\
& & \footnotesize{\quad\quad\quad\quad\enspace\, everyday with their kids then why aren't they seeing the signs? Oh wait, it's easier to blame everyone else} \\
\cmidrule(lr){3-3} & & \textbf{Target - } \footnotesize{\{Supporting | Refuting | Irrelevant | Discussing | Querying \}} \\
\midrule

\multirow{3.5}{*}{\textbf{FEVER}}  & \multirow{3.5}{*}{\footnotesize{Fact Verification}} & \textbf{Source - } \footnotesize{History of art includes architecture, dance, sculpture, music, painting, poetry literature, theatre, narrative, film, } \\
& & \footnotesize{\quad\quad\quad\quad\enspace\, photography and graphic arts.} \\
\cmidrule(lr){3-3} & & \textbf{Target - } \footnotesize{\{Supports | Refutes\}} \\
\midrule

\multirow{3.5}{*}{\textbf{\textsc{MultiPIT$_\texttt{Id}$}}}  & \multirow{3.5}{*}{\footnotesize{Paraphrase Identification}} & \textbf{Source - } \footnotesize{sentence1: well 160 people died in Bangladesh due to building collapse} \\
& & \footnotesize{\quad\quad\quad\quad\enspace\, sentence2: \#bangladesh Death toll climbs in Bangladesh building collapse} \\
\cmidrule(lr){3-3} & & \textbf{Target - } \footnotesize{\{Yes | No\}} \\
\midrule

\multirow{2}{*}{\textbf{\textsc{MultiPIT$_\texttt{Gen}$}}}  & \multirow{2}{*}{\footnotesize{Paraphrase Generation}} & \textbf{Source - } \footnotesize{President Obama will hold a press conference at 10:15 a.m. at the White House} \\
\cmidrule(lr){3-3} & & \textbf{Target - } \footnotesize{ President Obama will be taking questions from reporters at 10:15 am ET in the briefing room} \\
\midrule

\multirow{2}{*}{\textbf{\textsc{Natural Questions}}}  & \multirow{2}{*}{\footnotesize{Question Answering}} & \textbf{Source - } \footnotesize{Who is the first person who went to moon?} \\
\cmidrule(lr){3-3} & & \textbf{Target - } \footnotesize{\, Neil Alden Armstrong} \\
\bottomrule
\end{tabular}
\end{adjustbox}
\caption{Input-output examples for each task. 
}
\label{Table:input-output}
\end{table*}

\begin{table*}[ht!]
\centering
\begin{adjustbox}{width=1\textwidth}
\begin{tabular}{lccccccc}
\toprule
\textbf{Hyperparameters} & \textbf{WLP} & \textbf{\textsc{Stanceosaurus}} & \textbf{FEVER} & \textbf{\textsc{MultiPIT$_\texttt{Id}$}} & \textbf{\textsc{MultiPIT$_\texttt{Gen}$}} & \textbf{\textsc{Natural Questions}} \\
\midrule\midrule
Max Source Length & 128 & 128 & 128 & 64 & 32 & 32 \\
Max Target Length & 128 & 8 & 8 & 8 & 32 & 32 \\
Batch Size & 32 & 32 & 32 & 32 & 32 & 32 \\
Epochs & 50 (20) & 50 (20) & 50 (20) & 50 (20) & 50 (20) & 50 (20) \\
Learning Rate & 3e-5 & 3e-5 & 3e-5 & 3e-5 & 3e-5 & 1e-3 (3e-5) \\
\bottomrule
\end{tabular}
\end{adjustbox}
\caption{Hyperparameters used for training models. The numbers in () are used exceptionally for \texttt{T5-XXL} (i.e., teacher) fine-tuning.}
\label{Table:hyperparameters}
\end{table*}

\section{Detailed Settings and Hyperparameters}
\label{hyperparameters}
As described in \S \ref{settings}, we utilize \texttt{T5 v1.1} \cite{roberts-etal-2020-much} as a base model, because the original version of \texttt{T5} \cite{2020t5} was pre-trained using a combination of several supervised tasks as well as an unsupervised task. Since this work assumes that no supervised datasets are available, our fine-tuning strategies build upon \texttt{T5 v1.1} that was pre-trained in an unsupervised way only. For a question answering task, we exceptionally use the checkpoint additionally pre-trained using salient span masking (SSM), an unsupervised pre-training objective known to be helpful for open-domain question answering \cite{pmlr-v119-guu20a}, following \citet{roberts-etal-2020-much}.

Table \ref{Table:data_statistics} presents the dataset statistics and Table \ref{Table:hyperparameters} presents the hyperparameters used for training models for each task. We did not try to specifically tune the hyperparameters for each model for each task, taking into account the scenario considered by this study in which annotated data is highly limited. Moreover, in order to minimize factors other than the ones we consider for each setup, we fixed each parameter as much as possible unless significant problems were observed during training. Specifically, we chose the learning rate of 3e-5 (default in the Huggingface \cite{wolf2019huggingfaces} code base for question answering and seq2seq distillation), which we believe is not out of the ordinary, for all except for \textsc{Natural Questions} where we adopt 1e-3 when training \texttt{T5-Small} model as we observed the phenomenon that it was not being trained at all by looking at its training loss with 3e-5. We trained all models with 50 epochs except for a \texttt{T5-XXL} model where fewer epochs are assumed to be enough. We used the final batch size of 32 by leveraging the gradient accumulation (e.g., batch size of \{16, 8\} and gradient accumulation of \{2, 4\}) when necessary to meet VRAM constraints. We adopt (layer-wise) model parallelism that allows us to load a large model on multiple GPUs. Our reported results are based on a single run due to the high computational cost required by our empirical study. Despite this, a significant difference in performance was observed between the two strategies being compared.

\section{Unlabeled Data for Each Task}
\label{unlabeled_data}
For the distillation strategy, unlabeled data is essentially required to transfer a large model's knowledge into a small model. In this work, unlabeled data is literally referred to the data without the corresponding labels (i.e., only source inputs in Table \ref{Table:input-output}). We exploit only input sources (without annotations) in the existing datasets excluding ones that models are evaluated on. Plus, we collect additional unlabeled corpora for each dataset for an extensive study as follows: 

\paragraph{WLP} This dataset requires procedural text as an input source. We utilize large-scale \textsc{Procedure} corpus \cite{bai-etal-2021-pre} that contains diverse domains. We specifically use \textsc{ChemSyn}, chemical synthesis procedures in patents, for this study.

\paragraph{\textsc{Stanceosaurus}} The input source for this dataset consists of a claim from diverse fact-checking sites, a tweet relevant to the claim, and contextual information such as a reply or parent tweet if any. Following the methodology described in this work \cite{Zheng2022StanceosaurusCS}, we collected claims and corresponding tweets by anonymizing user information.

\paragraph{FEVER} Statements or claims are sufficient to be sources for this dataset. We leverage the synthetically generated claims in \citet{schuster-etal-2021-get}.

\paragraph{\textsc{MultiPIT}} The sources for this dataset are sentences written by Twitter users, which can be collected by following the method in \citet{Dou2022ImprovingLP}. For this work, we instead exploit sources of \textsc{MultiPIT$_\texttt{AUTO}$} \cite{Dou2022ImprovingLP} as unlabeled data, automatically collected recent datasets.

\paragraph{\textsc{Natural Questions}} The source simply consists of a question. Therefore, we make use of queries in MS MARCO \cite{DBLP:conf/nips/NguyenRSGTMD16}, where the queries are sampled from Bing's search logs.

\begin{figure}[h!]
    \centering
    \includegraphics[scale=0.27]{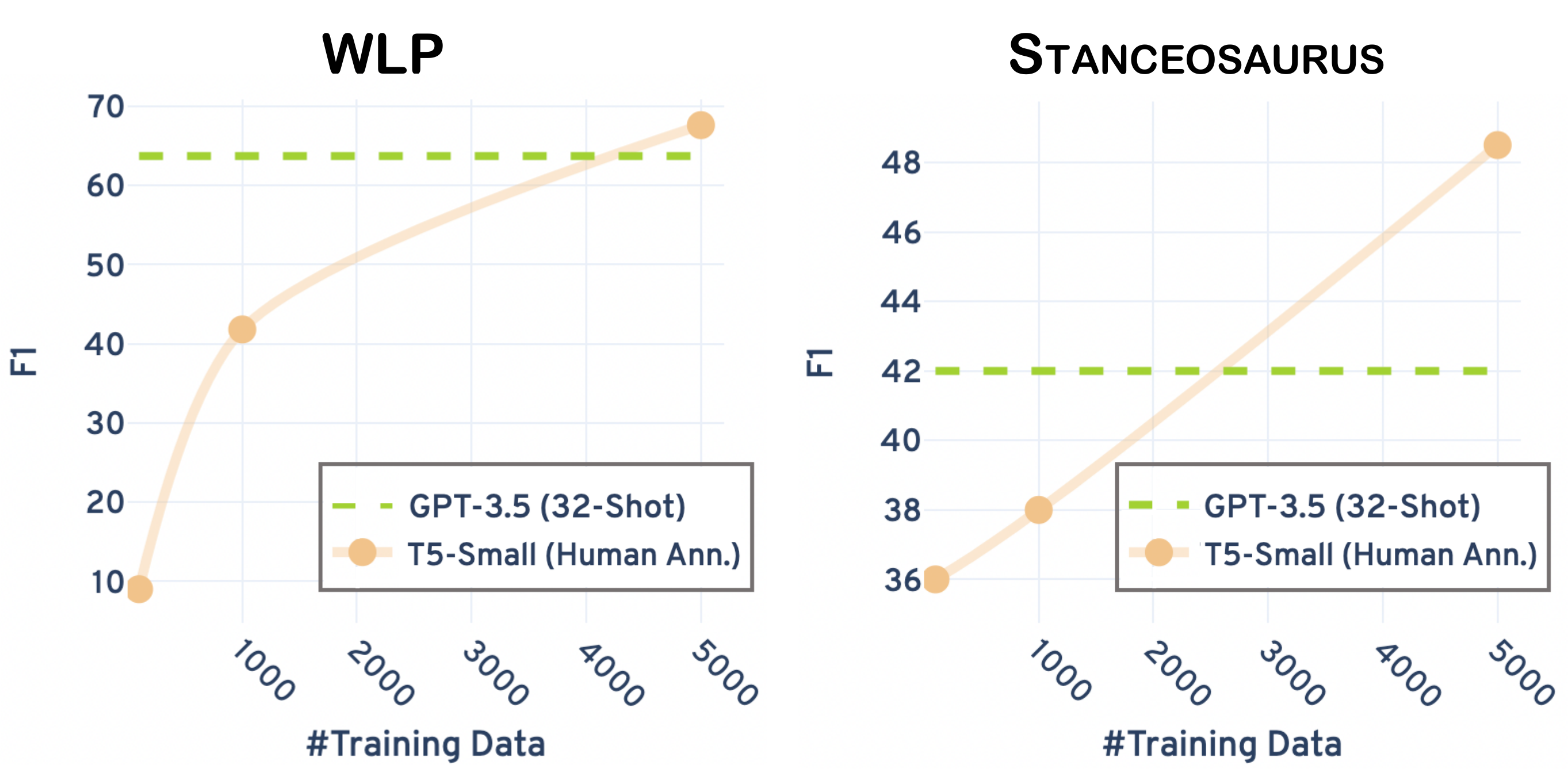}
    \caption{Preliminary results on 200 sampled test sets, comparing \texttt{GPT-3.5} 32-shot in-context learning against T5-Small with varying the size of training data.}
    \label{fig:gpt3.5_test}
\end{figure}

\section{Details of GPT-3.5 Annotation}
\label{gpt-3.5_details}
To annotate pseudo-labels using \texttt{GPT-3.5}, we make use of the strongest version, \texttt{text-davinci-003} with 32 training examples. Our input prompt consists of a task-specific instruction\footnote{For WLP, "Classify named entities into one of the following categories: \{Class 1, Class2, ...\}"}\footnote{For \textsc{Stanceosaurus}, "Classify the stance of a given tweet toward a given claim into one of the following categories: \{Class 1, Class2, ...\}"} and 32 in-context examples, and unlabeled input to annotate at the end. 
In order to reduce the high variance \cite{pmlr-v139-zhao21c, min-etal-2022-noisy}, we randomly sample and shuffle 32 in-context examples out of a 100 fixed training set for each annotation iteration.
In Figure \ref{fig:gpt3.5_test}, we present the performance of \texttt{GPT-3.5}'s 32-shot learning to see its quality and feasibility, and we find that it can be qualified as a cheap labeler to improve performances, especially for low-budget settings, as found in \citet{wang-etal-2021-want-reduce}.

Note that OpenAI\footnote{https://openai.com/api/pricing} API charges based on the number of tokens for input prompt plus model output: \texttt{\$0.02} per 1K tokens. Therefore, the \$ per label is calculated as \texttt{\$0.046} for WLP (2.3K tokens on average) and \texttt{\$0.073} for \textsc{Stanceosaurus} (3.65K tokens on average). 
Based on this, we annotate 4347 data for WLP and 2739 data for \textsc{Stanceosaurus} in total, using \texttt{\$200} assigned for each task.

\begin{table}[h!]
\centering
\begin{adjustbox}{width=0.48\textwidth}
\begin{tabular}{lcc}
\toprule
\textbf{Dataset} & \textbf{Exisiting Models} & \textbf{T5-XXL (Full)} \\
\midrule\midrule
\textbf{WLP} & 75.9 \cite{bai-etal-2021-pre} & 74.4  \\
\textbf{\textsc{Stanceosaurus}} & 61.0 \cite{Zheng2022StanceosaurusCS} & 63.3 [69.8] \\
\textbf{FEVER} & 78.9 \cite{petroni-etal-2021-kilt} & 82.1 \\
\textbf{\textsc{MultiPIT$_\texttt{Id}$}} & 91.4 \cite{Dou2022ImprovingLP} & 90.8 \\
\textbf{\textsc{MultiPIT$_\texttt{Gen}$}} & 77.8 \cite{Dou2022ImprovingLP} & 75.9 \\
\textbf{\textsc{Natural Questions}} & 35.2 \cite{roberts-etal-2020-much} & 31.3 [38.5] \\
\bottomrule
\end{tabular}
\end{adjustbox}
\caption{Resource-unconstrained performances of existing models and fully fine-tuned in-house T5-XXL for reference or upper bounds. Due to the use of different metrics, we also report macro F1 for \textsc{Stanceosaurus}, and the EM score for \textsc{Natural Questions}, along with the [micro F1] used in this work.}
\label{Table:perf_calibration}
\end{table}

\begin{table}[h!]
\centering
\begin{adjustbox}{width=0.48\textwidth}
\begin{tabular}{lcccc}
\toprule
\multirow{3}{*}{\textbf{Model}} & \multicolumn{2}{c}{\textbf{\textsc{Stanceosaurus}}} & \multicolumn{2}{c}{\textbf{FEVER}} \\
 & \texttt{\$664} & \texttt{\$2120} & \texttt{\$279} & \texttt{\$795} \\
 & (\textit{N=1K}) & (\textit{N=5K}) & (\textit{N=1K}) & (\textit{N=5K}) \\
\midrule\midrule
\textbf{\texttt{T5-Small \small{(Ann.)}}} & 44.7 & 50.3 & 49.8 & 68.9 \\
\textbf{\texttt{DistilBERT \small{(General Dist. + Ann.)}}} & 56.3 & 57.5 & 69.9 & 73.5 \\
\textbf{\texttt{BERT$_\textsc{Base}$ \small{(Ann.)}}} & 56.0 & 59.0 & 70.7 & 73.1 \\
\textbf{\texttt{T5-XXL $\Rightarrow$ T5-Small \small{(Dist.)}}} & 56.9 & 60.5 & 71.7 & 74.8 \\
\bottomrule
\end{tabular}
\end{adjustbox}
\caption{Results (\textit{N=5K}) of \texttt{Ann.}, general distillation (\textbf{\texttt{DistilBERT}} \cite{sanh2019distilbert}), and task-specific distillation on \textsc{Stanceosaurus} and FEVER. For \textbf{\texttt{DistilBERT}}, the computational cost for distillation in the pre-training phase is assumed to be \$0. The final model size is similar to each other ($\sim$60M) except for \textbf{\texttt{BERT$_\textsc{Base}$}} (110M). General (pre-training) distillation and task-specific (fine-tuning) distillation are complementary \cite{jiao2020tinybert}.}
\label{Table:distilbert}
\end{table}

\section{Additional Results}
\label{additional_results}

\paragraph{How well do off-the-shelf models perform for each task?} In Table \ref{Table:perf_calibration}, we provide the results of the largest \texttt{T5} model (11B) fined-tuned on full training data, along with relevant works' results in resource-rich settings. Those reported numbers can serve as upper bounds or references for calibrating the relative results produced in this work (i.e., resource-limited settings). Note that these should not be used for direct comparison due to various combinations of factors including model architectures, size, approaches, pre-training scheme, training data, and budgets.

\paragraph{What about general distillation?} While this work focuses on task-specific distillation, we also provide the result of general distillation (\texttt{DistilBERT} \cite{sanh2019distilbert}) in which a model is distilled during the pre-training phase to learn general language understanding capability before fine-tuning. To measure the total cost, the computational cost for distillation in the pre-training phase is assumed to be \texttt{\$0} (i.e., it is publicly available). In Table \ref{Table:distilbert}, we find that given the same budget, adding general distillation leads to more cost-efficient than the annotation strategy without distillation. In addition to this, it is important to note that intuitively, general distillation (pre-training) and task-specific (fine-tuning) distillation can be combined for the better, evidenced in \citet{jiao2020tinybert}. This further spotlights the cost-efficient aspect of distillation methods.

\begin{table*}[ht!]
\centering
\begin{adjustbox}{width=1\textwidth}
\begin{tabular}{lccrlrlrlrlrl}
\toprule
\multirow{4}{*}{\textbf{Task}} & \multirow{4}{*}{\textbf{\textit{\colorbox{labeled}{$N$}}} (\texttt{Initial \$})} & \multirow{4}{*}{\textbf{Strategy}}
& \multicolumn{8}{c}{\textbf{\texttt{Additional \$}}} \\
\cmidrule(lr){6-10}
    & & & \multicolumn{8}{c}{\texttt{Ann.} Performance \quad \footnotesize{(\textit{\colorbox{labeled}{\#Additional Data}})}} \\
    & & & \multicolumn{8}{c}{\texttt{Dist.} Performance \quad \footnotesize{(\textit{\colorbox{unlabeled}{GPU Hours / \#Unlabeled Data}})}} \\
\midrule\midrule

\multirow{4.3}{*}{\textbf{WLP}} & \multirow{4.5}{*}{\textit{\colorbox{labeled}{100}} (\texttt{\$26})}
 & & \multicolumn{2}{c}{\texttt{+\$0}} & \multicolumn{2}{c}{\texttt{+\$100}} & \multicolumn{2}{c}{\texttt{+\$200}} & \multicolumn{2}{c}{\texttt{+\$300}} \\
\cmidrule(lr){4-5}\cmidrule(lr){6-7}\cmidrule(lr){8-9}\cmidrule(lr){10-11}
 & & \texttt{T5-Small (Ann.)} & 
 \textbf{9.1} & \footnotesize{(\textit{\colorbox{labeled}{+0}})} &
 23.8 & \footnotesize{(\textit{\colorbox{labeled}{+384}})} & 
 37.1 & \footnotesize{(\textit{\colorbox{labeled}{+769}})} & 
 47.6 & \footnotesize{(\textit{\colorbox{labeled}{+1153}})} \\
 & & \texttt{T5-XXL [\textbf{48.8}] $\Rightarrow$ T5-Small (Dist.)} & 
 \multicolumn{2}{c}{\footnotesize{N/A}} &
 \textbf{49.5} & \footnotesize{(\textit{\colorbox{unlabeled}{54h / 22K}})} & 
 \textbf{49.5} & \footnotesize{(\textit{\colorbox{unlabeled}{107h / 45K}})} & 
 \textbf{49.9} & \footnotesize{(\textit{\colorbox{unlabeled}{160h / 68K}})} \\
\midrule\midrule

\multirow{4.3}{*}{\textbf{\textsc{Stanceosaurus}}} & \multirow{4.5}{*}{\textit{\colorbox{labeled}{100}} (\texttt{\$36})}
 & & \multicolumn{2}{c}{\texttt{+\$0}} & \multicolumn{2}{c}{\texttt{+\$100}} & \multicolumn{2}{c}{\texttt{+\$200}} & \multicolumn{2}{c}{\texttt{+\$300}} \\
\cmidrule(lr){4-5}\cmidrule(lr){6-7}\cmidrule(lr){8-9}\cmidrule(lr){10-11}
 & & \texttt{T5-Small (Ann.)} & 
 \textbf{35.2} & \footnotesize{(\textit{\colorbox{labeled}{+0}})} & 
 35.2 & \footnotesize{(\textit{\colorbox{labeled}{+274}})} & 
 45.2 & \footnotesize{(\textit{\colorbox{labeled}{+549}})} & 
 45.4 & \footnotesize{(\textit{\colorbox{labeled}{+824}})} \\
 & & \texttt{T5-XXL [\textbf{44.8}] $\Rightarrow$ T5-Small (Dist.)} & 
 \multicolumn{2}{c}{\footnotesize{N/A}} &
 \textbf{45.8} & \footnotesize{(\textit{\colorbox{unlabeled}{54h / 42K}})} & 
 \textbf{45.8} & \footnotesize{(\textit{\colorbox{unlabeled}{107h / 87K}})} & 
 \textbf{45.6} & \footnotesize{(\textit{\colorbox{unlabeled}{160h / 131K}})} \\
\midrule\midrule

\multirow{4.3}{*}{\textbf{\textsc{FEVER}}} & \multirow{4.5}{*}{\textit{\colorbox{labeled}{100}} (\texttt{\$13})}
 & & \multicolumn{2}{c}{\texttt{+\$0}} & \multicolumn{2}{c}{\texttt{+\$50}} & \multicolumn{2}{c}{\texttt{+\$100}} & \multicolumn{2}{c}{\texttt{+\$150}} \\
\cmidrule(lr){4-5}\cmidrule(lr){6-7}\cmidrule(lr){8-9}\cmidrule(lr){10-11}
 & & \texttt{T5-Small (Ann.)} & 
 \textbf{50.3} & \footnotesize{(\textit{\colorbox{labeled}{+0}})} & 
 49.3 & \footnotesize{(\textit{\colorbox{labeled}{+387}})} & 
 49.7 & \footnotesize{(\textit{\colorbox{labeled}{+775}})} & 
 49.7 & \footnotesize{(\textit{\colorbox{labeled}{+1162}})} \\
 & & \texttt{T5-XXL [\textbf{49.7}] $\Rightarrow$ T5-Small (Dist.)} & 
 \multicolumn{2}{c}{\footnotesize{N/A}} &
 \textbf{49.7} & \footnotesize{(\textit{\colorbox{unlabeled}{27h / 59K}})} & 
 49.7 & \footnotesize{(\textit{\colorbox{unlabeled}{54h / 123K}})} & 
 49.7 & \footnotesize{(\textit{\colorbox{unlabeled}{80h / 187K}})} \\
\midrule\midrule

\multirow{4.3}{*}{\textbf{\textsc{MultiPIT$_\texttt{Id}$}}} & \multirow{4.5}{*}{\textit{\colorbox{labeled}{100}} (\texttt{\$20})}
 & & \multicolumn{2}{c}{\texttt{+\$0}} & \multicolumn{2}{c}{\texttt{+\$100}} & \multicolumn{2}{c}{\texttt{+\$200}} & \multicolumn{2}{c}{\texttt{+\$300}} \\
\cmidrule(lr){4-5}\cmidrule(lr){6-7}\cmidrule(lr){8-9}\cmidrule(lr){10-11}
 & & \texttt{T5-Small (Ann.)} & 
 \textbf{46.9} & \footnotesize{(\textit{\colorbox{labeled}{+0}})} &
 53.1 & \footnotesize{(\textit{\colorbox{labeled}{+500}})} &
 53.1 & \footnotesize{(\textit{\colorbox{labeled}{+1000}})} &
 53.1 &  \footnotesize{(\textit{\colorbox{labeled}{+1500}})} \\
 & & \texttt{T5-XXL [\textbf{53.1}] $\Rightarrow$ T5-Small (Dist.)} & 
 \multicolumn{2}{c}{\footnotesize{N/A}} & 
 53.1 & \footnotesize{(\textit{\colorbox{unlabeled}{54h / 78K}})} & 
 53.1 & \footnotesize{(\textit{\colorbox{unlabeled}{107h / 159K}})} & 
 53.1 & \footnotesize{(\textit{\colorbox{unlabeled}{160h / 240K}})} \\
\midrule\midrule

\multirow{4.3}{*}{\textbf{\textsc{MultiPIT$_\texttt{Gen}$}}} & \multirow{4.5}{*}{\textit{\colorbox{labeled}{100}} (\texttt{\$37})}
 & & \multicolumn{2}{c}{\texttt{+\$0}} & \multicolumn{2}{c}{\texttt{+\$100}} & \multicolumn{2}{c}{\texttt{+\$200}} & \multicolumn{2}{c}{\texttt{+\$300}} \\
\cmidrule(lr){4-5}\cmidrule(lr){6-7}\cmidrule(lr){8-9}\cmidrule(lr){10-11}
 & & \texttt{T5-Small (Ann.)} & 
 \textbf{45.0} & \footnotesize{(\textit{\colorbox{labeled}{+0}})} &
 \textbf{53.1} & \footnotesize{(\textit{\colorbox{labeled}{+269}})} &
 \textbf{57.3} & \footnotesize{(\textit{\colorbox{labeled}{+539}})} &
 \textbf{59.5} &  \footnotesize{(\textit{\colorbox{labeled}{+808}})} \\
 & & \texttt{T5-XXL [\textbf{55.5}] $\Rightarrow$ T5-Small (Dist.)} & 
 \multicolumn{2}{c}{\footnotesize{N/A}} & 
 41.4 & \footnotesize{(\textit{\colorbox{unlabeled}{54h / 59K}})} & 
 40.6 & \footnotesize{(\textit{\colorbox{unlabeled}{107h / 120K}})} & 
 41.0 & \footnotesize{(\textit{\colorbox{unlabeled}{160h / 181K}})} \\
\midrule\midrule

\multirow{4.3}{*}{\textbf{\textsc{Natural Questions}}} & \multirow{4.5}{*}{\textit{\colorbox{labeled}{100}} (\texttt{\$13})}
 & & \multicolumn{2}{c}{\texttt{+\$0}} & \multicolumn{2}{c}{\texttt{+\$50}} & \multicolumn{2}{c}{\texttt{+\$100}} & \multicolumn{2}{c}{\texttt{+\$150}} \\
\cmidrule(lr){4-5}\cmidrule(lr){6-7}\cmidrule(lr){8-9}\cmidrule(lr){10-11}
 & & \texttt{T5-Small (Ann.)} & 
 \textbf{2.3} & \footnotesize{(\textit{\colorbox{labeled}{+0}})} & 
 3.3 & \footnotesize{(\textit{\colorbox{labeled}{+387}})} & 
 3.9 & \footnotesize{(\textit{\colorbox{labeled}{+775}})} & 
 4.2 & \footnotesize{(\textit{\colorbox{labeled}{+1162}})} \\
 & & \texttt{T5-XXL [\textbf{18.6}] $\Rightarrow$ T5-Small (Dist.)} & 
 \multicolumn{2}{c}{\footnotesize{N/A}} &
 \textbf{9.1} & \footnotesize{(\textit{\colorbox{unlabeled}{27h / 37K}})} & 
 \textbf{11.0} & \footnotesize{(\textit{\colorbox{unlabeled}{54h / 78K}})} & 
 \textbf{11.0} & \footnotesize{(\textit{\colorbox{unlabeled}{80h / 118K}})} \\
\bottomrule
\end{tabular}
\end{adjustbox}
\caption{Detailed results in a few-shot learning scenario (\textit{N}=100) to investigate the cost efficiency of a small model with more \texttt{data annotations (Ann.)} and \texttt{teacher [\textbf{performance}] $\Rightarrow$ student (small) distillation (Dist.)} on various NLP tasks. \colorbox{labeled}{$N$} indicates the number of starting data annotated with the corresponding (\texttt{initial \$}). (\textit{\colorbox{labeled}{\#Additional Data}}) refers to the number of annotated data additional to \colorbox{labeled}{$N$}, and (\textit{\colorbox{unlabeled}{GPU Hours}}) denotes the total GPU hours for fine-tuning the teacher model on \colorbox{labeled}{$N$} data, plus for the distillation into a small model using varying (\textit{\colorbox{unlabeled}{\#Unlabeled Data}}).}
\label{Table:main_small_n}
\end{table*}

\end{document}